\documentclass[10pt,onecolumn,letterpaper]{article}

\usepackage{cvpr}
\usepackage{times}
\usepackage{epsfig}
\usepackage{graphicx}
\usepackage{amsmath}
\usepackage{amssymb}
\usepackage{amsthm}
\usepackage{bm}
\usepackage[tight,footnotesize]{subfigure}

\usepackage[pagebackref=true,breaklinks=true,letterpaper=true,colorlinks,bookmarks=false]{hyperref}

\def\onedot{. }
\def\eg{\emph{e.g}\onedot} 
\def\ie{\emph{i.e}\onedot}

\def\etal{\emph{et al}\onedot}

\cvprfinalcopy 

\def\cvprPaperID{1427} 

\ifcvprfinal\fi
\begin{document}

\title{Substructure and Boundary Modeling for Continuous Action Recognition}

\author{
\;\;\;\; Zhaowen Wang\textsuperscript{\dag} \;\;\;\;\; Jinjun Wang\textsuperscript{\ddag} \;\;\;\;\; Jing Xiao\textsuperscript{\ddag} \\
\textsuperscript{\dag}Beckman Institute\\
University of Illinois at Urbana-Champaign\\
{\tt\small \{wang308, klin21, huang\}@ifp.uiuc.edu}
\and Kai-Hsiang Lin\textsuperscript{\dag} \;\;\;\;\; Thomas Huang\textsuperscript{\dag} \;\;\;\; \\
\textsuperscript{\ddag} ASD Group\\
Epson Research and Development, Inc\\
{\tt\small \{jwang, xiaoj\}@erd.epson.com}
}

\maketitle

\begin{abstract}
This paper introduces a probabilistic graphical model for continuous
action recognition with two novel components: substructure
transition model and discriminative boundary model. The first
component encodes the sparse and global temporal transition prior
between action primitives in state-space model to handle the large
spatial-temporal variations within an action class. The second component
enforces the action duration constraint in a discriminative way to
locate the transition boundaries between actions more accurately.
The two components are integrated into a unified graphical structure
to enable effective training and inference. Our comprehensive
experimental results on both public and in-house datasets
show that, with the capability to incorporate additional
information that had not been explicitly or efficiently modeled by
previous methods, our proposed algorithm achieved significantly
improved performance for continuous action recognition.
\end{abstract}


\section{Introduction} \label{sec_intro}

Understanding continuous human activities from videos, \ie
simultaneous segmentation and classification of actions, is a
fundamental yet challenging problem in computer vision. Many
existing works approach the problem using bottom-up
methods~\cite{Satkin:Modeling}, where segmentation is performed as
preprocessing to partition videos into coherent constituent parts,
and action recognition is then applied as an isolated classification step.
Although a rich literature exists for segmentation of time series,
such as change point detection~\cite{Harchaoui:Kernel}, periodicity
of cyclic events modeling~\cite{Cutler:Robust} and frame
clustering~\cite{Zhou:Aligned}, the methods tend to detect local
boundaries and lack the ability to incorporate global dynamics
of temporal events, which leads to under or over segmentation that
severely affects the recognition performance, especially for complex
actions with diversified local motion statistics~\cite{Hoai:Joint}.

\begin{figure}[t]
 \center
    \subfigure[]{\epsfig{file=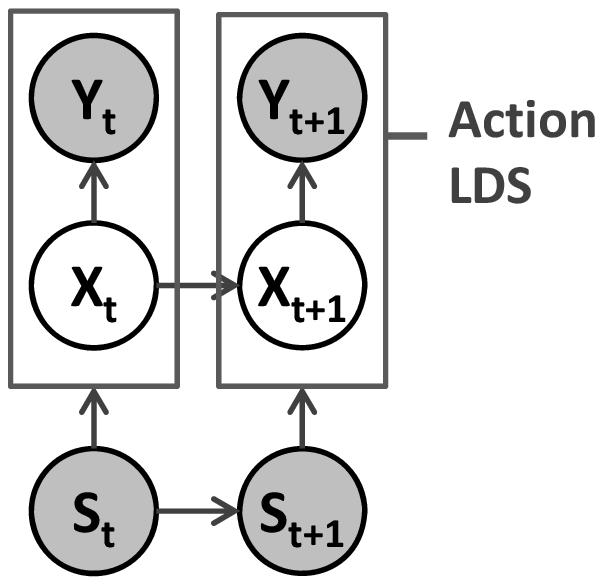, width=25mm,bbllx=52,bblly=52,bburx=224,bbury=218,clip=}}
    \hfil
    \subfigure[]{\epsfig{file=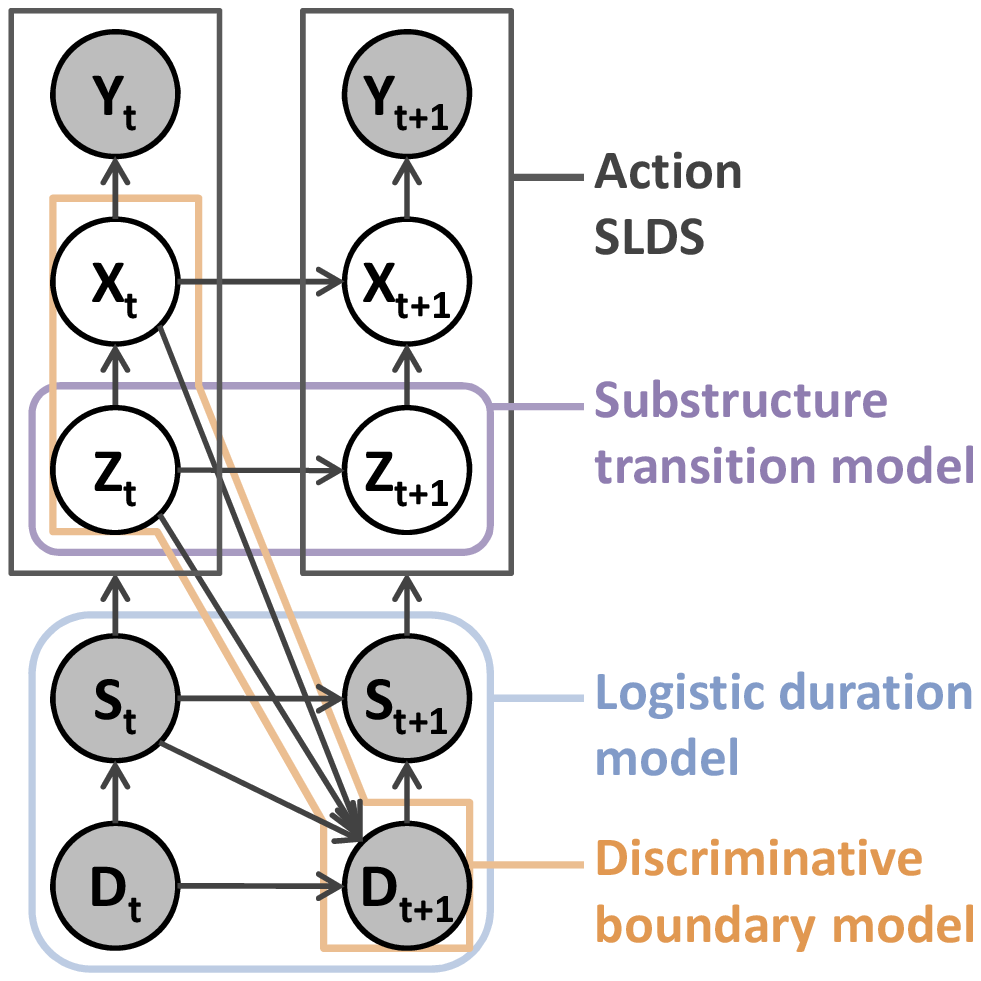, width=40mm, bbllx=70,bblly=50,bburx=350,bbury=330,clip=}}
  \caption{(a) Tradition SLDS model for continuous action recognition, where each action is
        represented by an LDS; (b) the structure of our proposed model, in which each action is represented
        by an SLDS with substructure transition, and the inter action transition is by controlled by discriminative boundary model.}
  \label{fig:overall_struct}
  \vspace{-0.15in}
\end{figure}

The limitation of the bottom-up approaches has been addressed by
performing concurrent top-down recognition using variants of Dynamic
Bayesian Network (DBN), where the dynamics of temporal events are
modeled as transitions in a latent~\cite{McCallum:Maximum,
Lafferty:Conditional} or partially observed state
space~\cite{Hoffken:Switching, Oh:Learning}. The technique has been
successfully used in speech recognition and natural language
processing, while the performance of existing DBN based approaches
for action recognition~\cite{Oh:Learning, Fox:Nonparametric,
Sminchisescu:Conditional, Sung:Human, Kjellstrom:Simultaneous,
Ning:Conditional} tends to be relatively lower~\cite{Hoai:Joint},
mostly due to the difficulty in interpreting the physical meaning of
latent states. Thus, it becomes difficult to impose additional prior
knowledge with clear physical meaning into an existing graphical
structure to further improve its performance.

To tackle the problem, in this paper, we show how two additional
sources of information with clear physical interpretations can be
considered in a general graphical structure for state-space model (SSM)
in Figure~\ref{fig:overall_struct}. Compared to a standard Switching
Linear Dynamic System (SLDS)~\cite{Oh:Learning} model in
Figure~\ref{fig:overall_struct}.(a), where $X$, $Y$ and $S$ are
respectively the hidden state, observation and label, the proposed model
in Figure~\ref{fig:overall_struct}.(b) is augmented with
two additional nodes, $Z$ and $D$, to describe the
substructure transition and duration statistics of actions:

\textbf{Substructure transition} Rather than a uniform motion type,
a real-world human action is usually characterized by a set of inhomogeneous units
with some instinct structure, which we call \emph{substructure}.
Action substructure arises from two factors: (1) the
hierarchical nature of human activity, where one action can be
temporally decomposed into a series of primitives with
spatial-temporal constraints; (2) the large variance of
action dynamics due to differences in kinematical property of subjects,
feedback from environment, or interaction with objects.
For the first factor,
Hoai~\etal~\cite{Hoai:Joint} used multi-class Support Vector Machine
(SVM) with Dynamic Programming to recognize coherent motion
constituent parts in an action; Liu \etal~\cite{Liu:Recognizing}
applied latent-SVM for temporal evolving of ``attributes'' in actions;
Sung \etal~\cite{Sung:Human} introduced a two-layer
Maximum Entropy Markov Models to recognize the correspondence
between sub-activities and human skeletal features.
For the second factor, considerations have been paid to
the substructure variance caused by subject-object interaction
using Connected Hierarchic Conditional Random Field (CRF)~\cite{Kjellstrom:Simultaneous},
and the substructure variance caused by pose using Latent Pose CRF~\cite{Ning:Conditional}.

In more general cases, Morency \etal presented the Latent Dynamic CRF
(LDCRF) algorithm by adding a ``latent-dynamic'' layer into CRF for
hidden substructure transition~\cite{Morency:Latent}. The
limitation of CRF as a discriminative method is that,
one single pseudo-likelihood score is estimated for
an entire sequence which is incapable to interpret the probability
of each individual frame.
To solve the problem, we instead design a generative model as
in Figure.\ref{fig:overall_struct}.(b), with extra hidden
node $Z$ gating the transition amongst a set of dynamic systems,
and the posterior for every action can be inferred strictly
under Bayesian framework for each frame.
The dimension of state space increases geometrically with an
extra hidden node, so we introduce effective transition prior constraints
in Section~\ref{sec:stm} to avoid over-fitting on a limited amount of training data.

\textbf{Duration model} The duration statistics of actions is
important in determining the boundary where one action
transits to another in continuous recognition tasks.
Duration model has been
widely adopted in Hidden Markov Model (HMM) based methods, such as
the explicit duration HMM~\cite{Ferguson:Variable} or more generally
the Hidden Semi Markov Model (HSMM)~\cite{Yu:Hidden}. Incorporating
duration model into SSM is more challenging than HMM because SSM has
continuous state space, and exact inference in SSM is usually
intractable~\cite{Lerner:inferencein}. Some works reported in this
line include Cemgil \etal\cite{Cemgil:generative} for music
transcription and Chib and Dueker~\cite{Chib:Non} for economics. Oh
\etal\cite{Sang:Parameterized} imposed the duration constraint at
the top level of SLDS and achieved improved performance for honeybee
behavior analysis~\cite{Oh:Learning}. In general, naive integration
of duration model into SSM is not effective, because duration
patterns vary significantly across visual data and limited training
samples may bias the model with incorrect duration patterns.

To address this problem, in Figure~\ref{fig:overall_struct}.(b) we
correlate duration node $D$ with the continuous hidden
state node $X$ and the substructure transition node $Z$ via
logistic regression as explained in Section~\ref{sec:abm}.
In this way, the proposed duration model
becomes more discriminative than conventional generative models,
and the data-driven boundary locating process can accommodate
more variation in duration length.

In summary, the major contribution of the paper is to incorporate
two additional models into a general SSM, namely the Substructure
Transition Model (STM) and the Discriminative Boundary Model (DBM).
We also design a Rao-Blackwellised particle filter for
efficient inference of proposed model in Section~\ref{sec:rbpf}.
Experiments in Section~\ref{sec:exp} demonstrate the
superior performance of our proposed system over several existing
state-of-the-arts in continuous action recognition.
Conclusion is drawn in Section~\ref{sec:con}.

\section{Substructure Transition Model}
\label{sec:stm}

\begin{figure}[t]
\center
    \subfigure[]{\epsfig{file=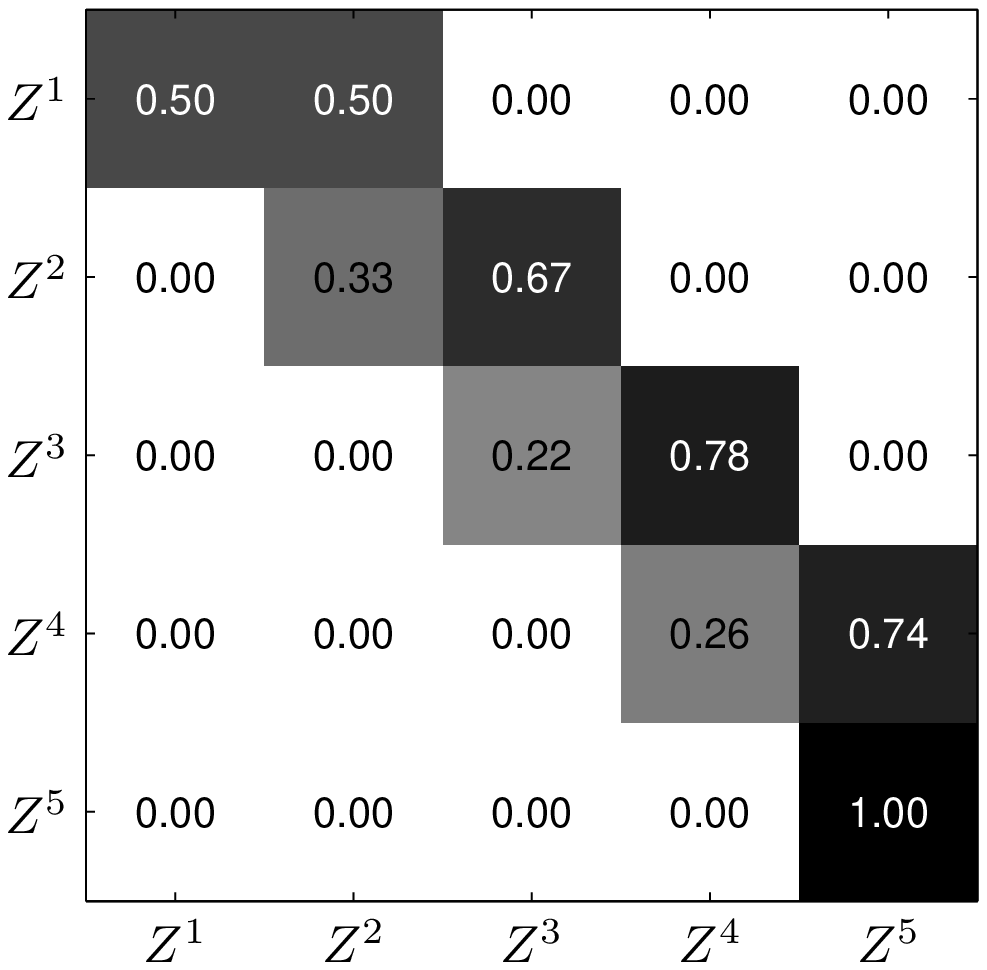, width=45mm,bbllx=161,bblly=254,bburx=462,bbury=546,clip}}
    \hfil
    \subfigure[]{\epsfig{file=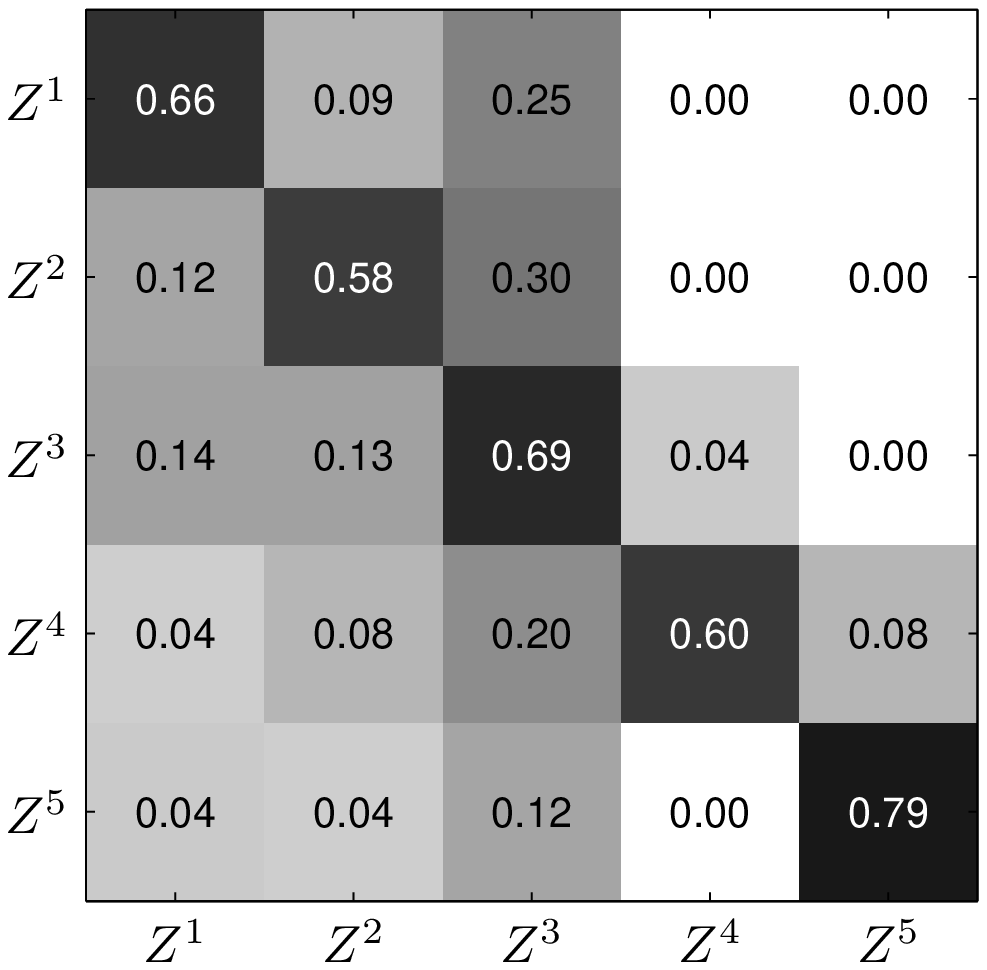, width=45mm,bbllx=161,bblly=254,bburx=462,bbury=546,clip}}
    \hfil
    \subfigure[]{\epsfig{file=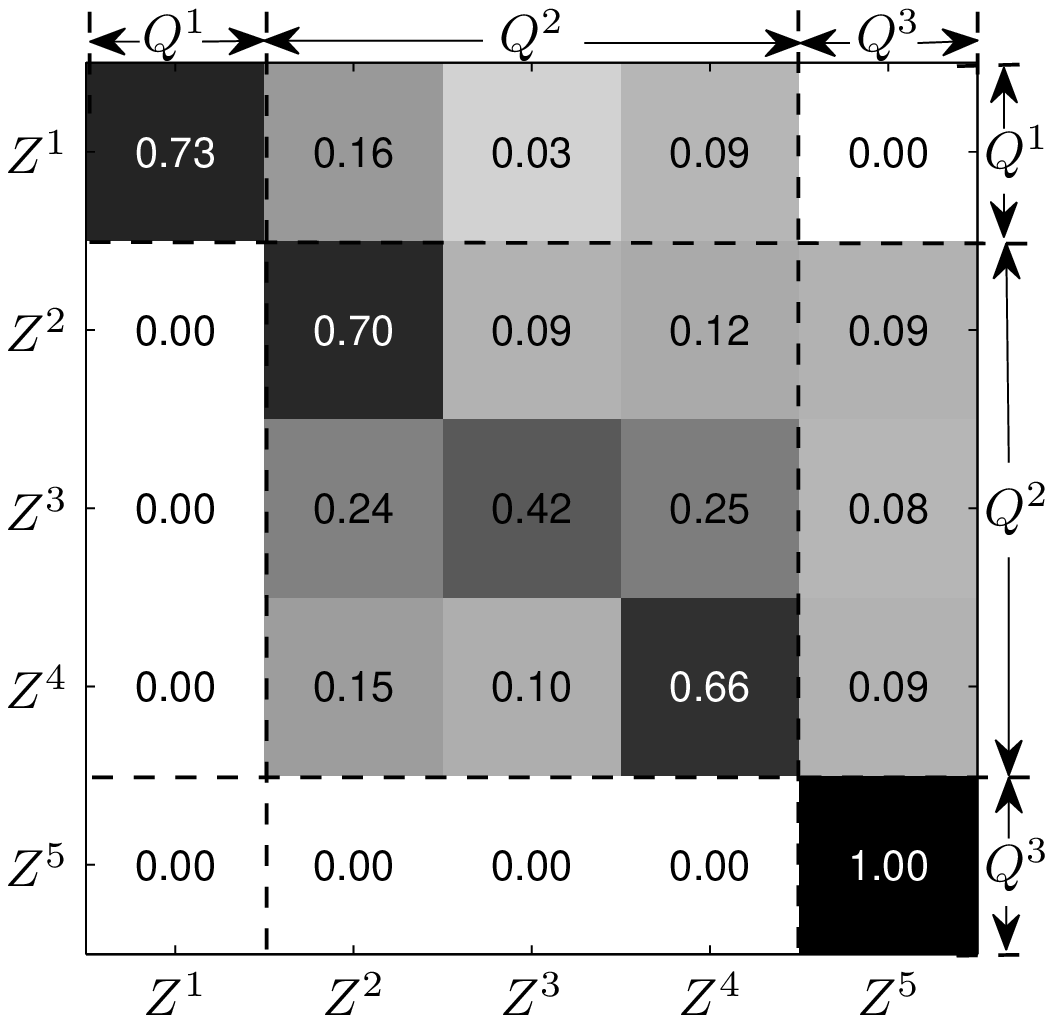, width=45mm,bbllx=161,bblly=254,bburx=462,bbury=546,clip}}
\caption{STM trained for action ``move-arm'' in stacking dataset using
        left-to-right (a), sparse (b), and block-wise sparse (c) constraints, with $N_Z=5$ and $N_Q=3$.
        STM in (c) captures global ordering and local details better than the other two.}
  \label{fig:stm_eg}
  \vspace{-0.1in}
\end{figure}

Linear Dynamic Systems (LDS) is the most commonly used SSM to
describe visual features of human motions. LDS is modeled by linear
Gaussian distributions:
\begin{equation}
\label{eq:lds_y}
    p(Y_t=\mathbf{y}_t|X_t=\mathbf{x}_t) = \mathcal{N}(\mathbf{y}_t; \mathbf{B} \mathbf{x}_t,  \mathbf{R})
\end{equation}
\begin{equation}
\label{eq:lds_x}
    p(X_{t+1}=\mathbf{x}_{t+1}|X_t=\mathbf{x}_t) = \mathcal{N}(\mathbf{x}_{t+1}; \mathbf{A} \mathbf{x}_t,  \mathbf{Q})
\end{equation}
where $Y_t$ is the observation at time $t$, $X_t$ is a latent state,
$\mathcal{N}(\mathbf{x}; \bm{\mu}, \mathbf{\Sigma})$ is
multivariate normal distribution of $\mathbf{x}$ with mean
$\bm{\mu}$ and covariance $\mathbf{\Sigma}$.
To consider multiple actions, SLDS~\cite{Oh:Learning} is formulated as a mixture of LDS's with
the switching among them controlled by action class $S_t$.
However, each LDS can only model an action with homogenous motion, ignoring
the complex substructure within the action.
We introduce a discrete hidden variable $Z_t \in \{1,...,N_Z\}$ to explicitly represent such information,
and the \emph{substructured} SSM can be stated as:
\begin{equation}
    p(Y_t=\mathbf{y}_t|X_t=\mathbf{x}_t, S_t^i, Z_t^j) = \mathcal{N}(\mathbf{y}_t; \mathbf{B}^{ij} \mathbf{x}_t,  \mathbf{R}^{ij})
\end{equation}
\begin{equation}
    p(X_{t+1}=\mathbf{x}_{t+1}|X_t=\mathbf{x}_t, S_{t+1}^i, Z_{t+1}^j) =
                \mathcal{N}(\mathbf{x}_{t+1}; \mathbf{A}^{ij} \mathbf{x}_t,  \mathbf{Q}^{ij})
\end{equation}
where $\mathbf{A}^{ij}$, $\mathbf{B}^{ij}$, $\mathbf{Q}^{ij}$, and
$\mathbf{R}^{ij}$ are the LDS parameters for the $j^{th}$ action primitive in the
substructure of $i^{th}$ action class. $\{Z_t\}$ is modeled as a
Markov chain and the transition probability is specified by
multinomial distribution:
\begin{equation}
\label{eq:trans_Z}
    p(Z_{t+1}^j|Z_t^i, S_{t+1}^k)=\theta_{ijk}
\end{equation}
In the following, the term STM may refer to either the transition matrix
in Eq. \eqref{eq:trans_Z} or the overall substructured SSM depending on its context.
Some examples of STM are given in Fig.~\ref{fig:stm_eg}, which are to be
explained in detail in the remainder of this section.

\subsection{Sparsity Constrained STM}

We use simplified notation $\mathbf{\Theta} = \{\theta_{ij}\}$ for
the STM within a single action.
In general, $\mathbf{\Theta}$ can be any matrix as long as each row of it is
a probability vector, which allows the substructure of action primitives
to be organized arbitrarily.
For most real-world human actions, however, there is a strong
temporal ordering associated with the primitive units. Such order
relationship can be vital to accurate action recognition,
since a different temporal ordering can define a totally different
action even if the composing primitive units are the same.
Moreover, if prior information of substructure is incorporated in model estimation,
the learning process can become more robust to noise and outliers.

There have been some attempts to characterize the order relationship of primitive units by
restricting the structure of transition matrix $\mathbf{\Theta}$.
For example, left-to-right HMM~\cite{Bakis:Continuous} is proposed to model sequential or cyclic ordering of primitive units,
and the corresponding $\mathbf{\Theta}$ has non-zero values only on or directly above diagonal,
as illustrated in Fig.~\ref{fig:stm_eg} (a).
Sequential ordering is a very strong assumption; to describe actions with more flexible temporal patterns,
people have resorted to switching HMM~\cite{Hoffken:Switching} or factorial HMM~\cite{Ghahramani:Factorial},
which model action variations with multiple sequential orderings.
All the works above assume the order relationship between action units are given \textit{a priori};
\ie, the number of non-zero entries in $\mathbf{\Theta}$ is small and their locations are all known.
In many cases, however, it is difficult to specify such information exactly,
and making a wrong assumption can bias the estimation of action model.
A more practical approach is to impose a sparse transition constraint while
leaving the discovery of exact order relationship to training phase.
Along this direction, negative Dirichlet distribution has been
proposed in \cite{Bicego:Sparseness} as a prior for each row $\bm{\theta}_i$
in $\mathbf{\Theta}$:
\begin{equation}
    p(\bm{\theta}_i) \propto \prod \limits_j \theta_{ij}^{-\alpha}
\end{equation}
where $\alpha$ is a pseudo count penalty. The MAP estimation of
parameter is
\begin{equation}
    \hat{\theta}_{ij} = \frac{\max(\xi_{ij}-\alpha, 0)}{\sum_t \max(\xi_{it}-\alpha, 0)}
\end{equation}
where $\xi_{ij}$ is the sufficient statistics of $(Z_t^i,
Z_{t+1}^j)$. When the number of transitions from $z^i$ to $z^j$ in
training data is less than $\alpha$, the probability $\theta_{ij}$
is set to zero. The sparsity enforced in this way often leads to
local transition patterns which might be actually caused by noise or
incomplete data, as shown in Fig.~\ref{fig:stm_eg} (b).
Also, the penalty term $\alpha$ introduces bias to
the proportion of non-zero transition probabilities, \ie
$\frac{\hat{\theta}_{ij}}{\hat{\theta}_{ik}} \neq
\frac{\xi_{ij}}{\xi_{ik}}$. This bias can be severe especially when
$\xi_{ij}$ is small.

\subsection{Block-wise Sparse STM}
\label{subsec:bsst}

As we have seen, the sequential order assumption about the transition between action units is too strong, while
the sparse prior on transition probability is biased and cannot globally regularize the STM.
Here we propose a block-wise sparse STM which can achieve tradeoff between model sparsity and
flexibility.
The idea is to divide an action into several stages and each stage comprises of a subset of action primitives.
The transition between stages is encouraged to be sequential but sparse,
such that the global action structure can be modeled. At the same time,
the action primitives within each stage can propagate freely from one to another
so that variation in action styles and parameters is also preserved.
Our stage-wise transition model is also favorable in regard of continuous action segmentation, since
the starting and terminating stages can be explicitly modeled to enhance discrimination
on action boundaries.


Formally, define discrete variable $Q_t \in \{1, ..., N_Q\}$ as the
current stage index of action, and assume a surjective mapping
$g(\cdot)$ is given which assigns each action primitive $Z_t$ to its
corresponding stage $Q_t$:
\begin{equation}
\label{eq:constraint_g}
    \left\{
    \begin{array}{ll}
    p(Q_t^q, Z_t^i) >0\,, & \textrm{if $g(i)=q$}\\
    p(Q_t^q, Z_t^i) =0\,. & \textrm{otherwise}
    \end{array}
    \vspace{-1mm}
\right.
\end{equation}

The choice of $g(\cdot)$ depends on the nature of action. Intuitively, we can assign more
action primitives to a stage with diversified motion patterns and less action primitives
to a stage with restricted pattern.
The joint dynamic transition distribution of $Q_t$ and $Z_t$ is defined as:
\begin{equation}
\label{eq:trans_QZ}
    p(Q_{t+1}, Z_{t+1} | Q_{t}, Z_{t})
     =  p(Q_{t+1} | Q_{t}) p(Z_{t+1} | Q_{t+1}, Z_{t})
\end{equation}
The second term of Eq. \eqref{eq:trans_QZ} specifies the transition
between action primitives, which we want to keep as flexible as
possible to model diversified local action patterns. The first term
captures the global structure between different action stages, and
therefore we impose an \emph{ordered} negative Dirichlet
distribution as its hyper-prior:
\begin{equation}
\label{eq:prior_phi}
    p(\mathbf{\Phi}) \propto \prod \limits_{q \neq r, q+1 \neq r} \phi_{qr}^{-\alpha}
\end{equation}
where $\mathbf{\Phi}=\{ \phi_{qr} \}$ is the stage transition
probability matrix, $\phi_{qr} = p(Q_{t+1}^r | Q_{t}^q)$, and
$\alpha$ is a constant for pseudo count penalty. The ordered
negative Dirichlet prior encodes both sequential order information
and sparsity constraint. It promotes statistically a global
transition path $Q^1 \rightarrow Q^2 \rightarrow ... \rightarrow
Q^{N_Q}$ which can be learned from training data rather than
heuristically defined as in left-to-right HMM
\cite{Bakis:Continuous}. An example of the resulting STM
is shown in Fig.~\ref{fig:stm_eg} (c).
Note that no in-coming{\slash}out-going
transition is encouraged for $Q^1${\slash}$Q^{N_Q}$, which stands
for starting{\slash}terminating stage. The identification of these
two special stages is helpful for segmenting continuous actions, as
will be discussed in Sec. \ref{sec:ddm}.

\subsection{Learning STM}

The MAP model estimation requires maximizing the product of
likelihood (\ref{eq:trans_QZ}) and prior (\ref{eq:prior_phi}) under
the constraint of (\ref{eq:constraint_g}).
There are two interdependent nodes, $Q$ and $Z$, involved in the optimization,
which make the problem complicated.
Fortunately, as shown in Appendix \ref{sec:app_blktrans}, Eq. \eqref{eq:trans_QZ}
can be replaced with the transition distribution of single variable $Z$ and
a constraint exists for the relationship between $\mathbf{\Theta}$ and $\mathbf{\Phi}$.
Therefore, the node $Q$ (and the associated parameter $\mathbf{\Phi}$) serves only for
conceptual purpose and can be eliminated in our model construction.
The MAP estimation can be converted to the following constrained
optimization problem:
\begin{eqnarray}
\label{eq:objfunc_blktrans}
    & \max \limits_{\mathbf{\Theta}} & \mathcal{L}(\mathbf{\Theta}) = \sum_{i, j} \xi_{ij} \log \theta_{ij}
                                    - \sum_{\substack{q \neq r \\ q+1 \neq r }} \alpha \log \phi_{qr} , \\
    & {\rm s.t.}  & \phi_{qr} = \Sigma_{j \in \mathcal{G}(r)} \theta_{ij}, \;\;\; i \in \mathcal{G}(q), \; \forall q, r \nonumber \\
    &             & \Sigma_{j} \theta_{ij} = 1, \;\;  \forall i   \;\;\; \;\;\; \;\;\;
                    \theta_{ij}\geq 0, \;\; \forall i, j \nonumber
\end{eqnarray}
where $\xi_{ij}$ is the sufficient statistics of $(Z_t^i, Z_{t+1}^j)$,
$\mathcal{G}(q) \triangleq \{ i | g(i)=q \}$,
and $\{\phi_{qr}\}$ are just auxiliary variables.
The KKT (Karush-Kuhn-Tucker) conditions for optimal solution $\mathbf{\hat{\Theta}}$ are:
\begin{eqnarray}
  \frac{\xi_{ij}}{\hat{\theta}_{ij}} - \lambda_{i, g(j)} + \gamma_i - \mu_{ij}  & = & 0, \;\; \forall i, j \nonumber \\
  -\frac{\alpha_{qr}}{\hat{\phi}_{qr}} + \sum_{i \in \mathcal{G}(q)} \lambda_{ir} & = & 0,  \;\; \forall q, r \nonumber \\
  \mu_{ij} \geq 0, \;\; \mu_{ij}\hat{\theta}_{ij}  & = & 0, \;\; \forall i, j \nonumber
\end{eqnarray}
where $\lambda_{ir}$, $\gamma_i$, and $\mu_{ij}$ are constant multipliers;
$\alpha_{qr}$ is equal to $\alpha$ if $q \neq r$ or $q+1 \neq r$, and $0$ otherwise.
Solving the equation set as in Appendix \ref{sec:app_thetahat} gives
the MAP parameter estimation:
\begin{eqnarray}
\label{eq:theta_hat}
    \hat{\theta}_{ij} &=& \hat{\phi}_{g(i), g(j)} \frac{\xi_{ij}}{\sum_{j' \in \mathcal{G}(g(j))} \xi_{ij'}} \\
    \hat{\phi}_{qr} &=&
        \frac{\max(\sum_{i \in \mathcal{G}(q), j \in \mathcal{G}(r)} \xi_{ij} - \alpha_{qr}, 0)}
        {\sum_{r'} \max(\sum_{i \in \mathcal{G}(q), j \in \mathcal{G}(r')} \xi_{ij} - \alpha_{qr'}, 0)}
     \nonumber
\end{eqnarray}
As we can see, the resultant transition
matrix $\hat{\mathbf{\Theta}}$ is a block-wise sparse matrix, which
can characterize both the global structure and local detail of
action dynamics. Also, within each block (stage), there is no bias
in $\hat{\theta}_{ij}$.


\section{Discriminative Boundary Model}
\label{sec:abm}

\subsection{Logistic Duration Model}
\label{sec:ldm}

It is straightforward to use a Markov chain to model the transition
of action $S_t$ where $p(S^j_{t+1}|S^i_t)=a_{ij}$. The duration
information of the $i^{th}$ action is naively incorporated into its
self-transition probability $a_{ii}$, which leads to an
action duration model with exponential distribution:
\begin{equation}
    p(dur_i=\tau) = a_{ii}^{\tau-1} (1-a_{ii}), \;\; \tau=1, 2, 3 ... \nonumber
\end{equation}
Unfortunately, only a limited number of real-life events have an
exponentially diminishing duration. Inaccurate duration modeling can
severely affect our ability to segment consecutive actions and
identify their boundaries.

Non-exponential duration distribution can be implemented with
duration-dependent transition matrix, such as the one used in
HSMM~\cite{Yu:Hidden}. Fitting a transition matrix for each epoch within
the maximum length of duration is often impossible given a limited
number of training sequences, even when parameter hyperprior
such as hierarchical Dirichlet distribution~\cite{Wang:Event}
is used to restrict model freedom.
Parametric duration distributions such as
gamma~\cite{Levinson:Continuously}
and Gaussian~\cite{Yoshimura:Duration} provide a more compact way
to represent duration and show good performance in signal synthesis.
However, they are less useful in inference because the corresponding
transition probability is not easy to evaluate.

Here a new logistic duration model is proposed to address the above
limitations. We introduce a variable $D_t$ to represent the length of
time current action has been lasting. $\{D_t\}$ is a counting
process starting from $1$, and the beginning of a new action
is triggered whenever it is reset to $1$:
\begin{equation}
\label{eq:pS_trans} p(S_{t+1}^j | S_t^i, D^d_{t+1}) =
    \left\{
        \begin{array}{ll}
        \delta(j-i), & \textrm{if $d>1$}\\
        a_{ij}, & \textrm{if $d=1$}
        \end{array}
    \right.
    \vspace{-1mm}
\end{equation}
where $a_{ij}$ is the probability of transiting from previous action
$i$ to new action $j$. Notice that the same type of action can be
repeated if we have $a_{ii}>0$.

\begin{figure}[t]
\centerline{
    \subfigure[]{\includegraphics[width=60mm]{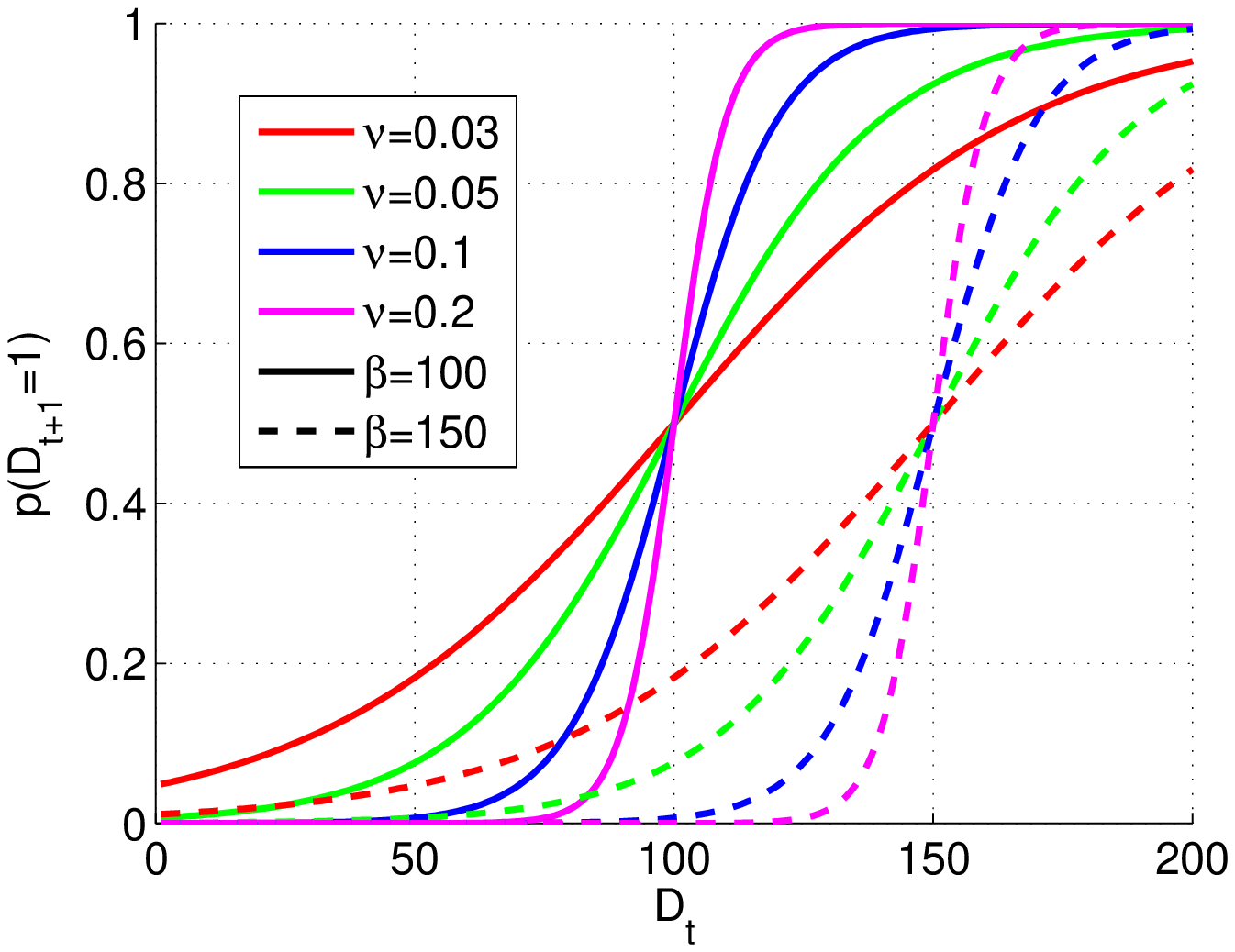}}
    \hfil
    \subfigure[]{\includegraphics[width=60mm]{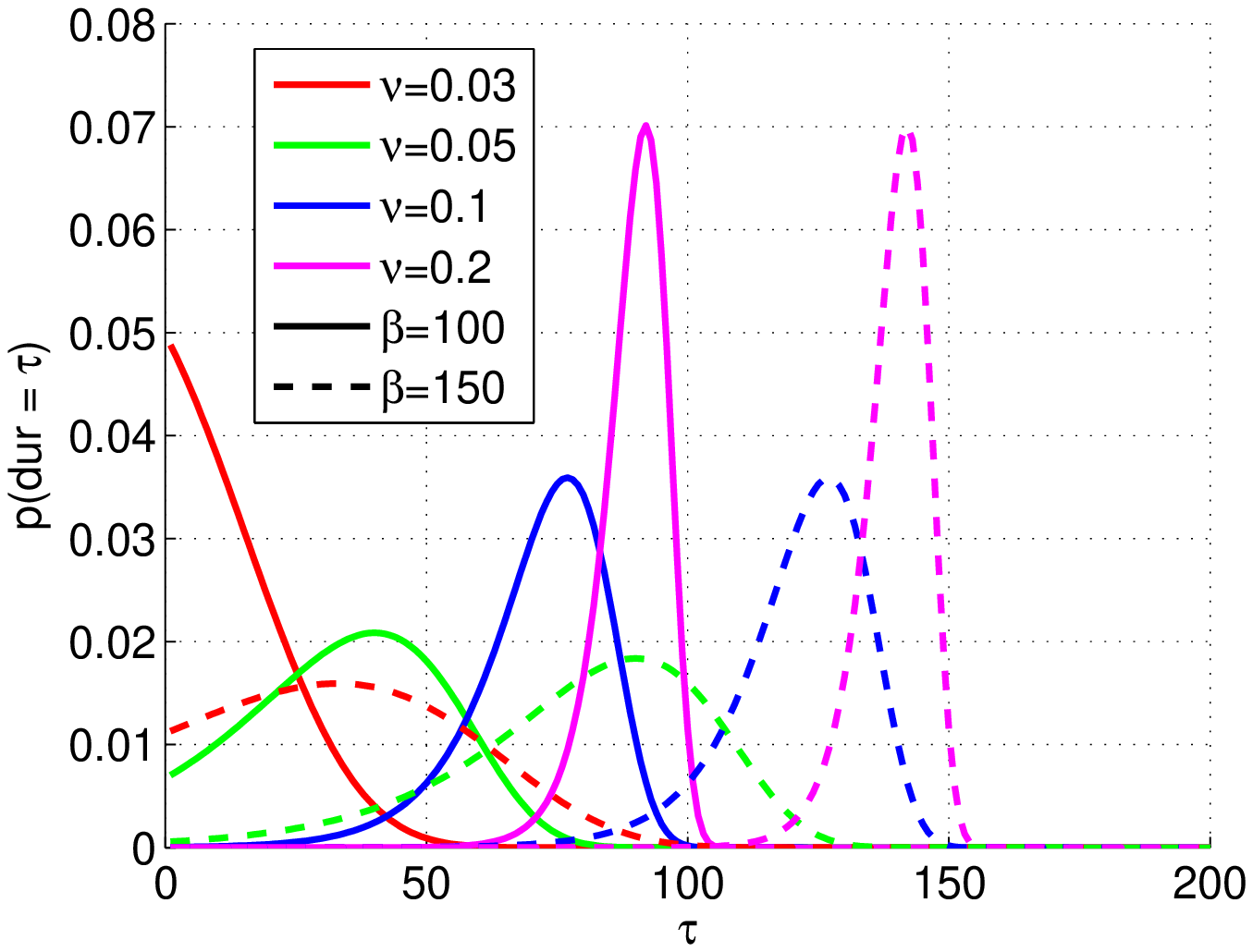}} }
\caption{(a) Resetting probability $p(D_{t+1}=1|D_t, S_t)$ and (b)
duration distribution for logistic duration model.
        Plotted with different color/line style for different $\nu$/$\beta$.}
  \label{fig:logistic_duration_curve}
  \vspace{-0.1in}
\end{figure}

Instead of modeling action duration distribution directly, we model
the transition distribution of $D_t$ as a logistic function of
its previous value:
\begin{equation}
\label{eq:pD_trans}
    p(D_{t+1}^c|S_{t}^i, D_{t}^d) = \frac{e^{\nu_i(d-\beta_i)}\delta(c-1)+\delta(c-d-1)}{1+e^{\nu_i(d-\beta_i)}}
\end{equation}
\begin{equation}
\label{eq:pD_init}
    p(D_1^c) = \delta(c-1)
\end{equation}
where $\nu_i$ and $\beta_i$ are positive logistic regression
weights. Eq. \eqref{eq:pD_trans} immediately leads to the duration
distribution for action class $i$:
\begin{equation}
\label{eq:pDur_logistic}
    p(dur_i=\tau)=\prod_{d=1}^{\tau} \frac{1}{1+e^{\nu_i(d-\beta_i)}} \times e^{\nu_i(\tau-\beta_i)}
\end{equation}
Fig.~\ref{fig:logistic_duration_curve} (a) shows how the resetting probability
of $D_{t+1}$ changes as a function of $D_{t}$ with different
parameter sets, and the corresponding duration distributions
are plotted in (b). The increasing probability of
transiting to a new action leads to a peaked duration distribution,
with center and width controlled by $\beta_i$ and $\nu_i$, respectively.
Our logistic duration model
can be easily extended to represent multiple-mode durations if
double logistic function~\cite{Lipovetsky:Double} is used.

\subsection{Discriminative Boundary Model}
\label{sec:ddm}

The logistic duration model can be integrated with STM
by stacking the two layers of nodes ($D$-$S$ and $Z$-$X$-$Y$) together.
The resultant generative model, however, is unable to utilize
contextual information for accurate action boundary segmentation.
Discriminative graphic models, such as MEMM~\cite{McCallum:Maximum}
and CRF~\cite{Lafferty:Conditional}, are generally more powerful in
such classification problem except that they ignore data likelihood
or suffer from label bias problem.

To integrate discriminating power into our action boundary model
and at the same time keep the generative nature of the action model
itself, we construct DBM by further augmenting the duration distribution
(which triggers action boundary) with the contextual information
from latent states $X$ and $Z$:
\begin{equation}
\label{eq:pD_trans_disc}
    p(D_{t+1}^1|S_{t}^i, D_{t}^d, X^{\mathbf{x}}_t, Z^j_t) =
    \frac{e^{\nu_i (d-\beta_i) + \bm{\omega}_{ij}^T \mathbf{x}}}
         {1+e^{\nu_i (d-\beta_i) + \bm{\omega}_{ij}^T \mathbf{x}}}
\end{equation}
where $\nu_i$, $\beta_{i}$ have the same meaning as in
Eq. \eqref{eq:pD_trans}, and $\bm{\omega}_{ij}$ are the additional
logistic regression coefficients. When $\bm{\omega}_{ij}^T
\mathbf{x}=0$, no information can be learned from $X_t$ and $Z_t$,
and the DBM reduces to a generative one as
Eq. \eqref{eq:pD_trans}. A similar logistic function has been
employed in augmented SLDS \cite{Barber:Expectation}, where the main
motivation is to distinguish between transitions to different states
based on latent variable. Our DBM is specifically
designed for locating the boundary between contiguous actions. It
relies on both real valued and categorical inputs.

As constrained by the STM in Subsection~\ref{subsec:bsst},
each action is only likely to terminate in stage $N_Q$.
Therefore, $D_{t+1}$ can be reset to $1$
only when the current action in this terminating stage, and we can
modify Eq. \eqref{eq:pD_trans_disc} as: \vspace{-1mm}
\begin{equation}
        \addtolength{\tabcolsep}{-3pt}
\label{eq:pD_trans_disc_term}
    p(D_{t+1}^1|S_{t}^i, D_{t}^d, X^{\mathbf{x}}_t, Z^j_t) =
    \left\{
        \begin{tabular}{ll}
        \textrm{Eq. \eqref{eq:pD_trans_disc}}, & \textrm{$g(j)=N_Q$}\\
        0, & \textrm{otherwise}
        \end{tabular}
    \right.
\end{equation}
In this way, the number of parameters is greatly reduced and the label
unbalance problem is also ameliorated.
Now, the construction of our action model for continuous recognition
has been completed, with the overall structure shown in Figure~\ref{fig:overall_struct} (b).

\subsection{Learning DBM}


To learn the parameters $\nu$, $\beta$ and $\bm{\omega}$, we use
coordinate descent method to iterate between $\{\nu, \beta\}$ and
$\bm{\omega}$.
For $\nu$ and $\beta$, given a set of $N$ training sequences with class
labels $\{\mathbf{S}^{(n)}\}_{n=1...N}$, we can easily obtain the values
for all duration nodes $\{\mathbf{D}^{(n)}\}_{n=1...N}$
according to Eq. \eqref{eq:pS_trans}-\eqref{eq:pD_init}.
Then fitting the parameters $\nu$ and $\beta$ is equivalent
to performing logistic regression with input-output pairs
$(D^{(n)}_t, \delta(S^{(n)}_{t+1}-S^{(n)}_t))$.
The action transition probability $\{a_{ij}\}$ can be obtained trivially.

To estimate $\bm{\omega}_{ij}$, let $\{ T^{(n)} \}_{n=1...N}$ be our
training set, where each data sample
$T^{(n)}$ is a realization of all the nodes involved in
Eq. \eqref{eq:pD_trans_disc} at a particular time instance $t^{(n)}$
and $S_{t^{(n)}}=i$. Since $X_{t^{(n)}}$ and $Z_{t^{(n)}}$ are
hidden variables, their posterior $p(Z^j_{t^{(n)}}|\cdot)=p_Z^{(n)}$
and $p(X_{t^{(n)}}^{\mathbf{x}}|Z^j_{t^{(n)}},
\cdot)=\mathcal{N}(\mathbf{x}; \bm{\mu}^{(n)},
\mathbf{\Sigma}^{(n)})$ are first inferred from single action STM,
where the posterior of $X_{t^{(n)}}$ is approximated by a
Gaussian. The estimation of $\bm{\omega}_{ij}$ is obtained by
maximizing the expected log likelihood:
\begin{align}
\label{eq:max_omega}
      &  \max_{\bm{\omega}_{ij}} \sum_{n}
    \mathrm{E}_{p(X^{\mathbf{x}}_{t^{(n)}}, Z^j_{t^{(n)}}|\cdot)} \left[ \log l^{(n)}(\mathbf{x}, \bm{\omega}_{ij}) \right] \\
     = & \max_{\bm{\omega}_{ij}} \sum_{n}
    p_Z^{(n)} \int_{\mathbf{x}}  \log l^{(n)}(\mathbf{x}, \bm{\omega}_{ij}) \mathcal{N}(\mathbf{x}; \bm{\mu}^{(n)}, \mathbf{\Sigma}^{(n)})
    d \mathbf{x} \nonumber
\end{align}
where
\begin{equation}
    l^{(n)}(\mathbf{x}, \bm{\omega}) =
        \frac{e^{(c^{(n)} + \bm{\omega}^T \mathbf{x}){b^{(n)}}}}
         {1+e^{c^{(n)} + \bm{\omega}^T \mathbf{x}}}
\end{equation}
and $b^{(n)}=p(D_{t^{(n)}+1}=1)$, $c^{(n)}=\nu_i
(D_{t^{(n)}}-\beta_i)$.
The integral in Eq. \eqref{eq:max_omega} cannot be solved
analytically. Instead, we use unscented transform \cite{Julier:new}
to approximate the integral with the average over a set of sigma
points of $\mathcal{N}(\mathbf{x}; \bm{\mu}^{(n)},
\mathbf{\Sigma}^{(n)})$:
\begin{equation}
    \mathbf{x}^{(n)}_k  =
    \left\{
        \begin{array}{ll}
        \bm{\mu}^{(n)}, & k=0 \\
        \bm{\mu}^{(n)}+(\sqrt{M \mathbf{\Sigma}^{(n)}})_k, & k=1,...,M \\
        \bm{\mu}^{(n)}-(\sqrt{M \mathbf{\Sigma}^{(n)}})_{k-M}, & k=M+1,...,2M \\
        \end{array}
    \right.
    \nonumber
\end{equation}
where $M$ is the dimension of $\mathbf{x}$,
$(\sqrt{\mathbf{\Sigma}})_k$ is the $k^{th}$ column of the matrix
square root of $\mathbf{\Sigma}$. Therefore, Eq. \eqref{eq:max_omega}
converts to a weighted logistic regression problem with features
$\{\mathbf{x}^{(n)}_k\}$, labels $\{b^{(n)}\}$ and weights
$\{p_Z^{(n)}/(2M+1)\}$.

\section{Inference with Rao-Blackwellised Particle Filter}
\label{sec:rbpf}

In testing, given an observation sequence $\mathbf{y}_{1:T}$, we
want to find the MAP action labels $\hat{S}_{1:T}$ and the boundaries defined
by $\hat{D}_{1:T}$; we are also interested in the style of actions which
can be revealed from $\hat{Z}_{1:T}$. To obtain these MAP estimates,
we are required to find the posterior $p( S_{1:T}, D_{1:T},
Z_{1:T}|\mathbf{y}_{1:T})$, which is a non-trivial job given the
complicated hierarchy and nonlinearity of our model. We propose to
use particle filtering~\cite{Arulampalam02} for online inference due
to its capability in non-linear scenario. Moreover, the latent
variable $X_t$ can be marginalized by
Rao-Blackwellisation~\cite{Doucet00}. In this way, the computation
of particle filtering is significantly reduced since Monte Carlo
sampling is only conducted in the joint space of $(S_t, D_t, Z_t)$,
which has a much lower dimension and a highly compact support (note
the sparse transition probability between these variables).

Formally, we decompose the posterior distribution of all the hidden nodes at time $t$ as
\begin{equation}
   p(S_t, D_t, Z_t, X_t | \mathbf{y}_{1:t}) 
 =   p(S_t, D_t, Z_t | \mathbf{y}_{1:t}) p(X_t | S_t, D_t, Z_t, \mathbf{y}_{1:t})
\end{equation}
where the second term can be evaluated analytically because $X_t$ depends on other variables
through linear and Gaussian relations.
In Rao-Blackwellised particle filter~\cite{Khan04}, a set of $N_P$ samples
$\{(s^{(n)}_t, d^{(n)}_t, z^{(n)}_t)\}_{n=1}^{N_P}$ and the associated weights $\{w^{(n)}_t\}_{n=1}^{N_P}$
are used to approximate the intractable first term, while the second term is represented by
$\{\chi_t^{(n)}\}_{n=1}^{N_P}$, which are analytical distributions of $X_t$ conditioned on
corresponding samples:
\begin{equation}
    \chi_t^{(n)}(X_t) \triangleq p(X_t | s^{(n)}_t, d^{(n)}_t, z^{(n)}_t, \mathbf{y}_{1:t})
\end{equation}
In our model, $\chi_t^{(n)}(X_t) = \mathcal{N}(X_t; \hat{\mathbf{x}}^{(n)}_t,  \mathbf{P}^{(n)}_t)$
is a Gaussian distribution. Thus, the posterior can be represented as
\begin{equation}
   p(S_t, D_t, Z_t, X_t | \mathbf{y}_{1:t}) 
 \approx  \sum \limits_{n=1}^{N_P} w^{(n)}_t \delta_{S_t}(s^{(n)}_t) \delta_{D_t}(d^{(n)}_t)
           \delta_{Z_t}(z^{(n)}_t) \chi_t^{(n)}(X_t)
\end{equation}
where the approximation error approaches to zero as $N_P$ increases to infinite.

Given the samples $\{s^{(n)}_{t-1}, d^{(n)}_{t-1}, z^{(n)}_{t-1}, \chi_{t-1}^{(n)}\}$
and weights $\{w^{(n)}_{t-1}\}$ at time $t-1$, the posterior of $(S_t, D_t, Z_t)$ at time $t$ is:
\begin{equation}
\label{eq:rbpf_sdz}
          p(S_t, D_t, Z_t | \mathbf{y}_{1:t}) \propto
    \sum_n w^{(n)}_{t-1} p(S_t| D_t, s^{(n)}_{t-1}) 
     \times p(Z_t | S_t, D_t, z^{(n)}_{t-1}) \mathcal{L}^{(n)}_{t}(S_t, D_t,Z_t) \nonumber
\end{equation}
where
\begin{equation}
\label{eq:int_norm_logit}
    \mathcal{L}^{(n)}_{t}(S_t, D_t, Z_t) = \int p(\mathbf{y}_t | \mathbf{x}_{t-1}, S_t, Z_t)
                    \chi_{t-1}^{(n)}(\mathbf{x}_{t-1})  
    \times p(D_t|s^{(n)}_{t-1}, d^{(n)}_{t-1}, z^{(n)}_{t-1}, \mathbf{x}_{t-1}) d\mathbf{x}_{t-1}
\end{equation}
The detailed derivation is shown in Appendix \ref{sec:app_rbpfsdz}.
It is hard to draw sample from Eq. \eqref{eq:int_norm_logit}. Instead, we draw new samples
$(s^{(n)}_{t}, d^{(n)}_{t}, z^{(n)}_{t})$ from a proposal density defined as:
\begin{equation}
    q(S_t, D_t, Z_t | \cdot) =  p(S_t| D_t, s^{(n)}_{t-1}) p(Z_t | S_t, D_t, z^{(n)}_{t-1}) 
                         \times p(D_t | s^{(n)}_{t-1}, d^{(n)}_{t-1}, z^{(n)}_{t-1}, \hat{\mathbf{x}}_{t-1}^{(n)} )
\end{equation}
The new sample weights are then updated as follows:
\begin{equation}
    w^{(n)}_t \propto w^{(n)}_{t-1}
            \frac{\mathcal{L}^{(n)}_{t}(s^{(n)}_t, d^{(n)}_t, z^{(n)}_t)}
                 {p(d^{(n)}_t | s^{(n)}_{t-1}, d^{(n)}_{t-1}, z^{(n)}_{t-1}, \hat{\mathbf{x}}_{t-1}^{(n)} )}
\end{equation}
$\mathcal{L}^{(n)}_{t}(\cdot)$ is essentially the integral of a
Gaussian function with a logistic function. Although not solvable
analytically, it can be well approximated by a re-parameterized
logistic function according to \cite{maragakis08}.
Details on how to evaluate $\mathcal{L}^{(n)}_{t}(\cdot)$
can be found in Appendix \ref{sec:app_logit}.

Once we get $s^{(n)}_{t}$ and $z^{(n)}_{t}$,
$\chi_t^{(n)}$ is simply updated by Kalman filter.
Re-sampling and normalization procedures are applied after all the
samples are updated as in~\cite{Doucet00}.

\section{Experimental Results}
\label{sec:exp}

Our model is tested on four datasets for continuous action recognition.
In all the experiments, we have used parameters $N_Q=3$, $N_Z=5$, $N_P=200$.
First STM is trained independently for each action
using the segmented sequences in training set; then DBM is learned
from the inferred terminal stage of each sequence. The overall learning procedure
follows EM paradigm where the beginning and terminating stages are initially
set as the first and last $15\%$ of each sequence, and
the initial action primitives are obtained from K-means clustering.
In testing, after the online inference using particle filter, we further
adjust each action boundary using an off-line inference within a local neighborhood
of length $40$ centered at the initial boundary; in this way, the locally ``full'' posterior
in Sec. \ref{sec:rbpf} is considered.
We evaluate the recognition performance by per-frame accuracy.
Contribution from each model component (STM and DBM) is analyzed separately.

\subsection{Public Dataset} \label{subsec_pubset}

The first public dataset used is the IXMAS
dataset~\cite{Weinland:Free}. The dataset contains 11 actions, each
performed 3 times by 10 actors. The videos are acquired
using 5 synchronized cameras from different angles, and the actors
freely changed their orientation in acquisition.
We calculate dense optical flow in the silhouette area
of each subject, from which Locality-constrained
Linear Coding features (LLC)\footnote{implementation from author's
website}~\cite{Jinjun:LLC} are extracted as the observation
in each frame.
We have used 32 codewords and $4 \times 4$, $2 \times 2$ and $1 \times 1$
spatial pyramid~\cite{Lazebnik:Beyond}.
Table~\ref{tab_exp_ixmas} reports the continuous action recognition
results, in comparison with SLDS\footnote{implementation based on
BNT from http://code.google.com/p/bnt/}~\cite{Oh:Learning},
CRF\footnotemark[1]~\cite{Lafferty:Conditional} and
LDCRF\footnotemark[1]~\cite{Morency:Latent}.
Our proposed model (and each of its components) achieves a recognition accuracy
higher than all the other methods by more than $10\%$.

\begin{table}[th]
  \caption{Continuous action recognition for IXMAS dataset}\label{tab_exp_ixmas}
  \centering \small
  \begin{tabular}{|c|c|c||c|c|c|}
    \hline
    SLDS & CRF & LDCRF & STM & DBM & STM+DBM\\
    \hline
    53.6\% & 60.6\% & 57.8\% & 70.2\% & 74.5\% & \textbf{76.5}\%\\
    \hline
  \end{tabular}
\end{table}

The second public dataset used is the CMU MoCap
dataset~\footnote{http://mocap.cs.cmu.edu/}. For comparison purpose,
we report the results from the complete subset of subject 86. The
subset has 14 sequences with 122 actions in 8 category.
Quaternion feature is derived from the raw MoCap data as our observation
for inference. Table~\ref{tab_exp_cmu} lists the continuous action
recognition results, in comparison with the same set of benchmark
techniques as in the first experiment, as well
as~\cite{Ozay:Sequential, Raptis:Spike}. Similarly, results from
this experiment demonstrated the superior performance of our method.
It is interesting to note that, in Table~\ref{tab_exp_cmu}, the
frame-level accuracy by using DBM alone is a little higher than its combination
with STM. This is because there's only one subject in this
experiment and no significant variation in substructure is presented
in each action type, so temporal duration plays a more important
role in recognition. Nevertheless, the result attained by STM+DBM
is superior than all benchmark methods.

\subsection{In-house dataset} \label{subsec_house}

\begin{figure}[t]
  \center \small
  \addtolength{\tabcolsep}{-7pt}
  \begin{tabular}{cccccccc}
      \epsfig{file=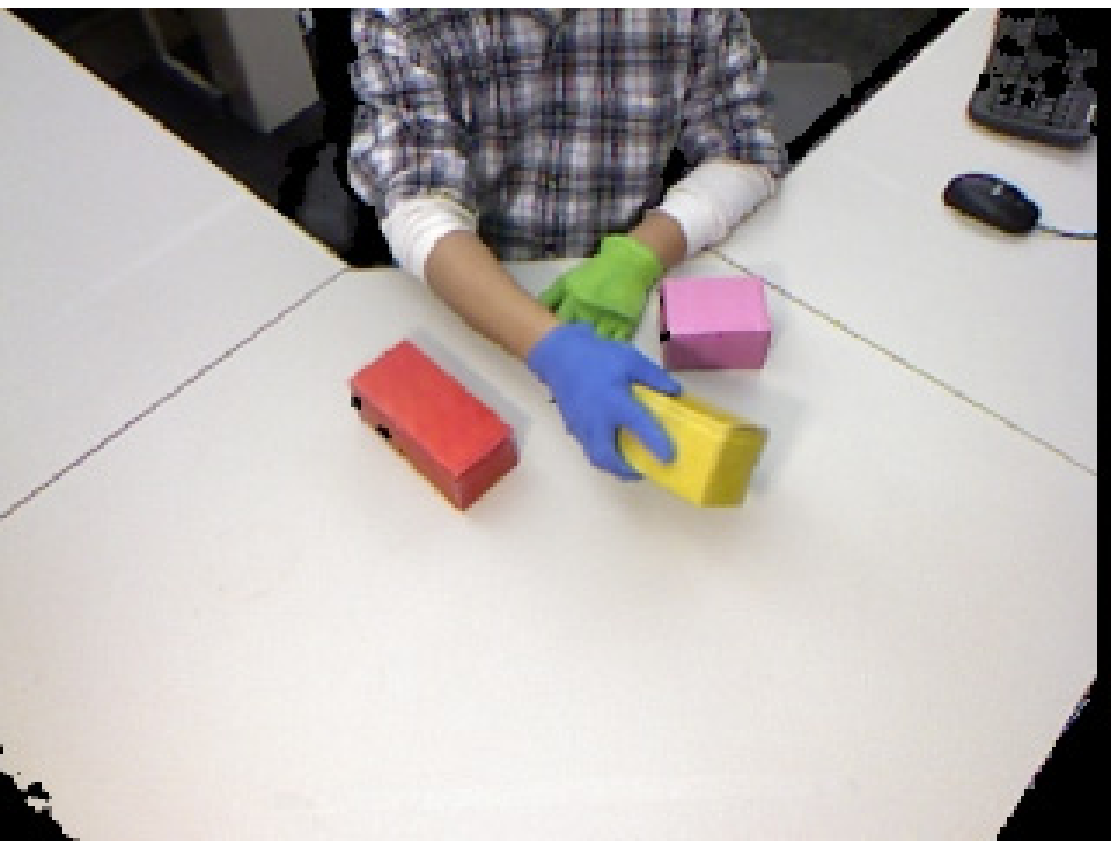, width=0.8in, height=0.5in} &
      \epsfig{file=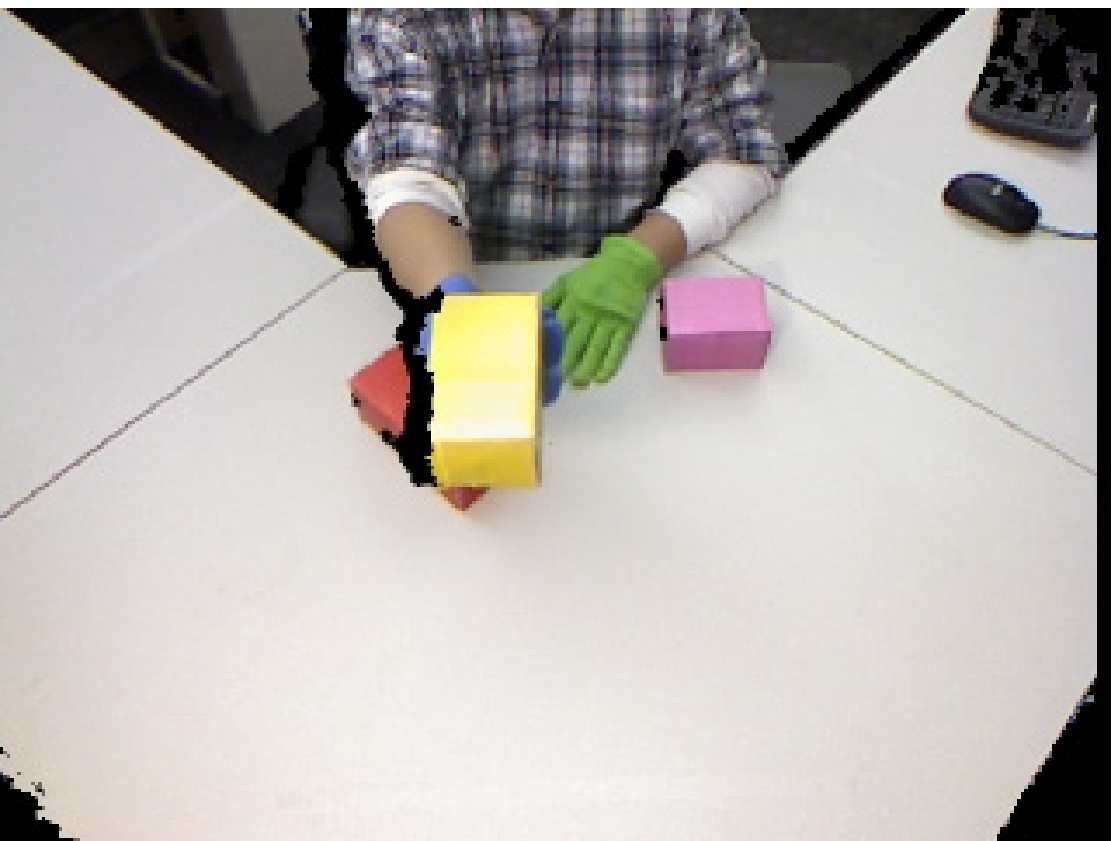, width=0.8in, height=0.5in} &
      \epsfig{file=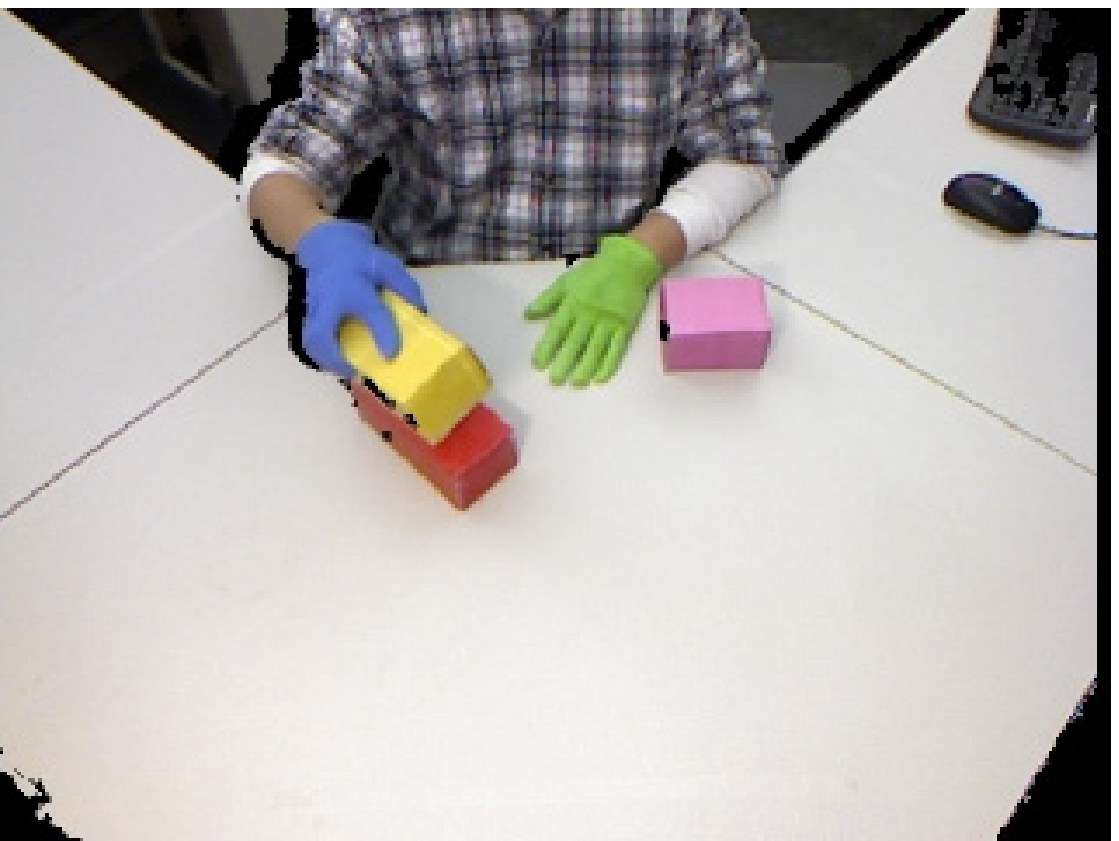, width=0.8in, height=0.5in} &
      \epsfig{file=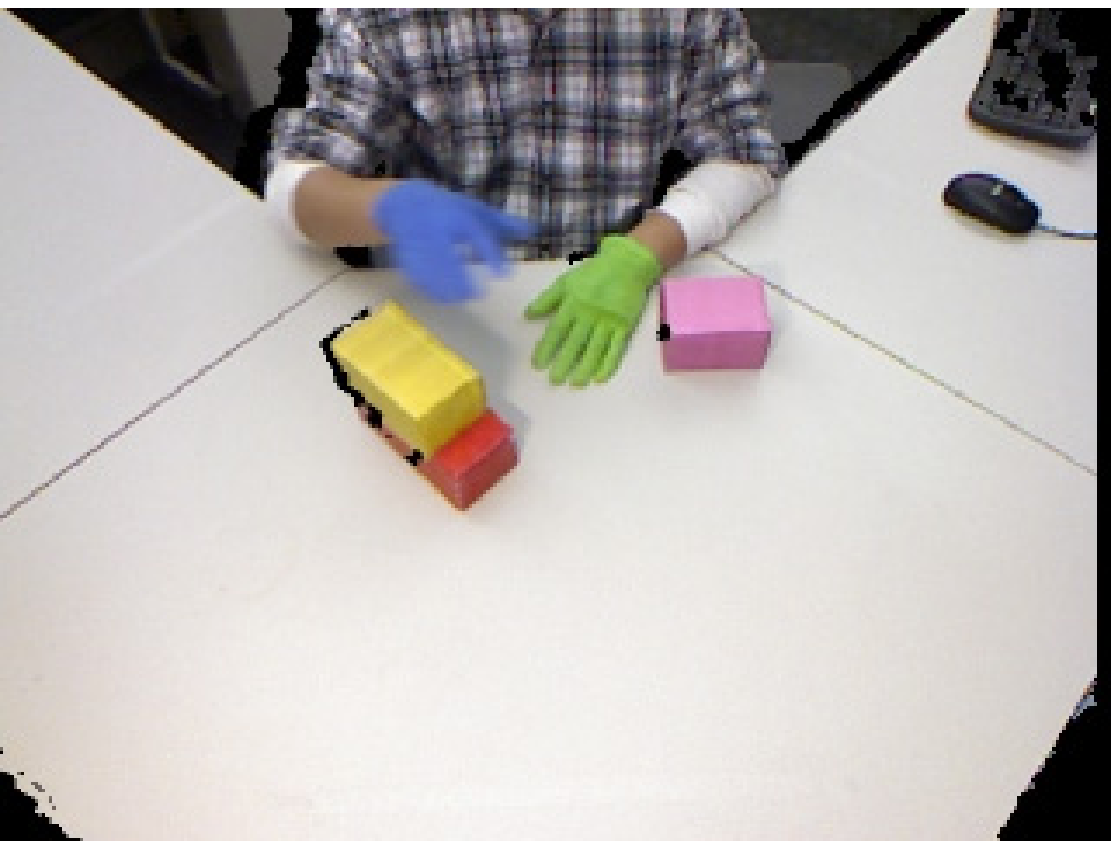, width=0.8in, height=0.5in} &
      \epsfig{file=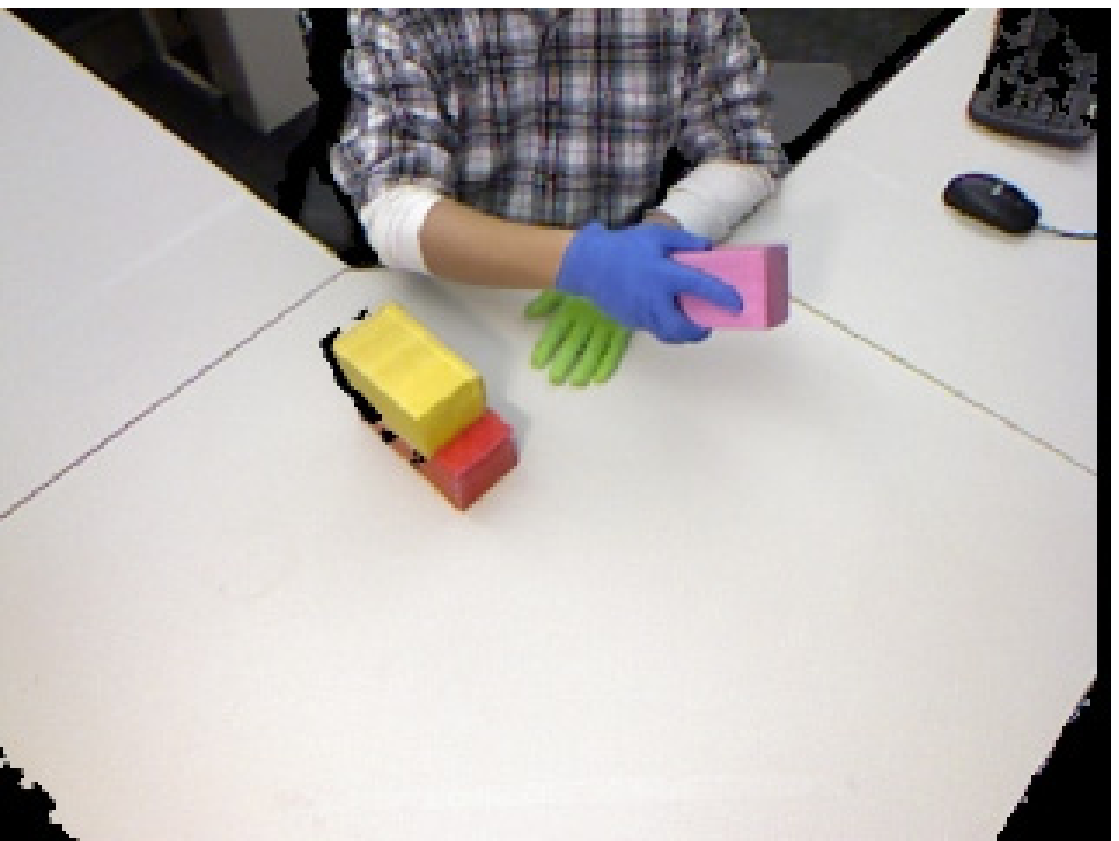, width=0.8in, height=0.5in} &
      \epsfig{file=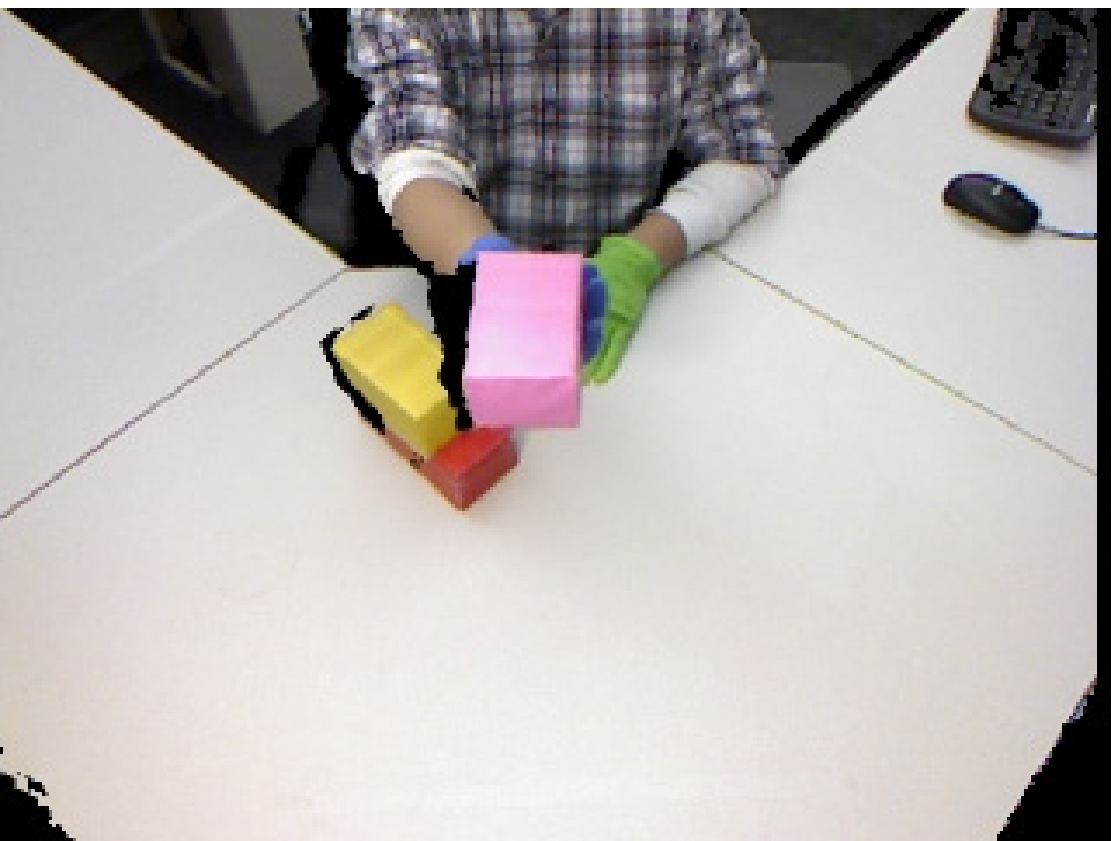, width=0.8in, height=0.5in} &
      \epsfig{file=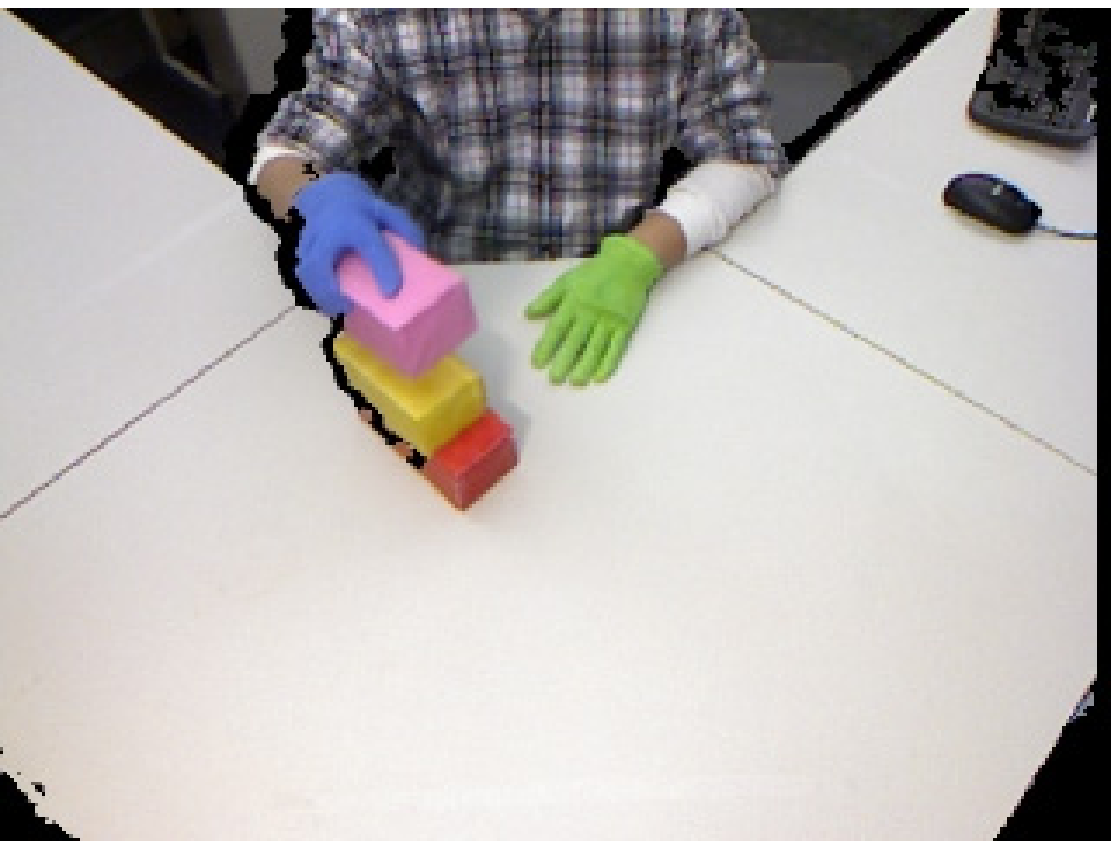, width=0.8in, height=0.5in} &
      \epsfig{file=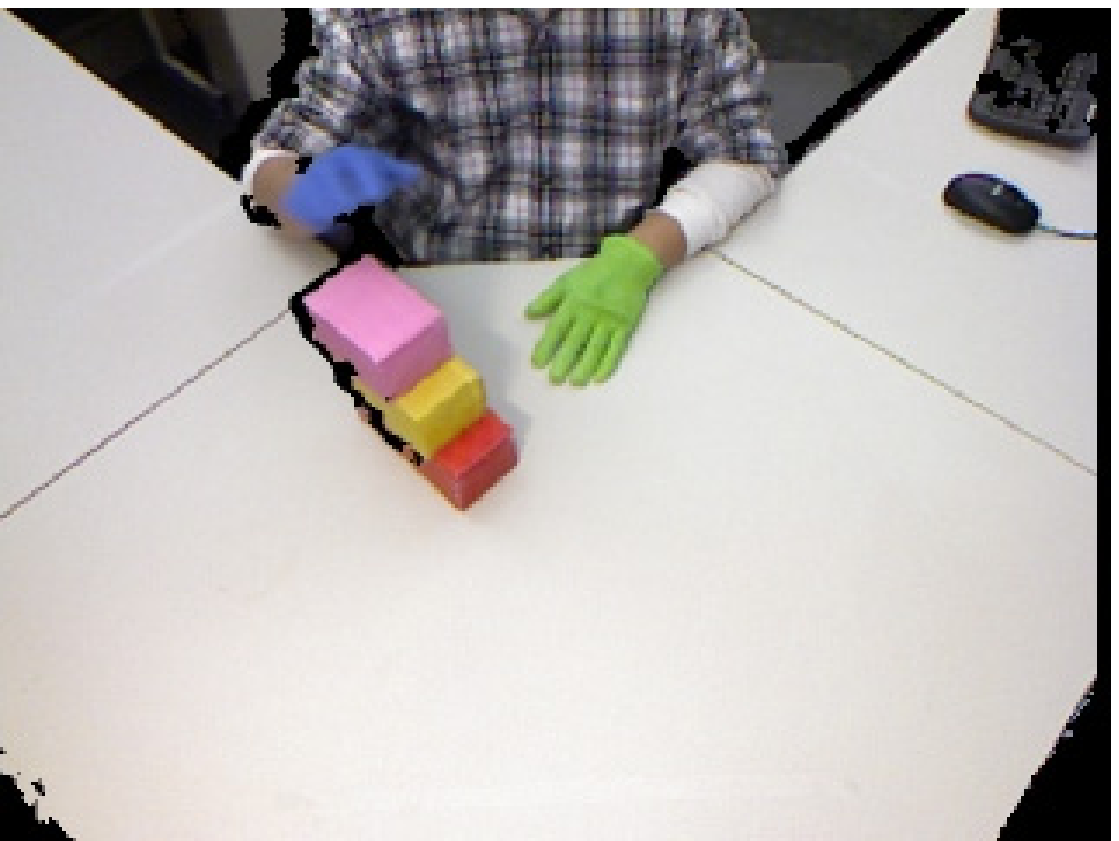, width=0.8in, height=0.5in} \\
      \epsfig{file=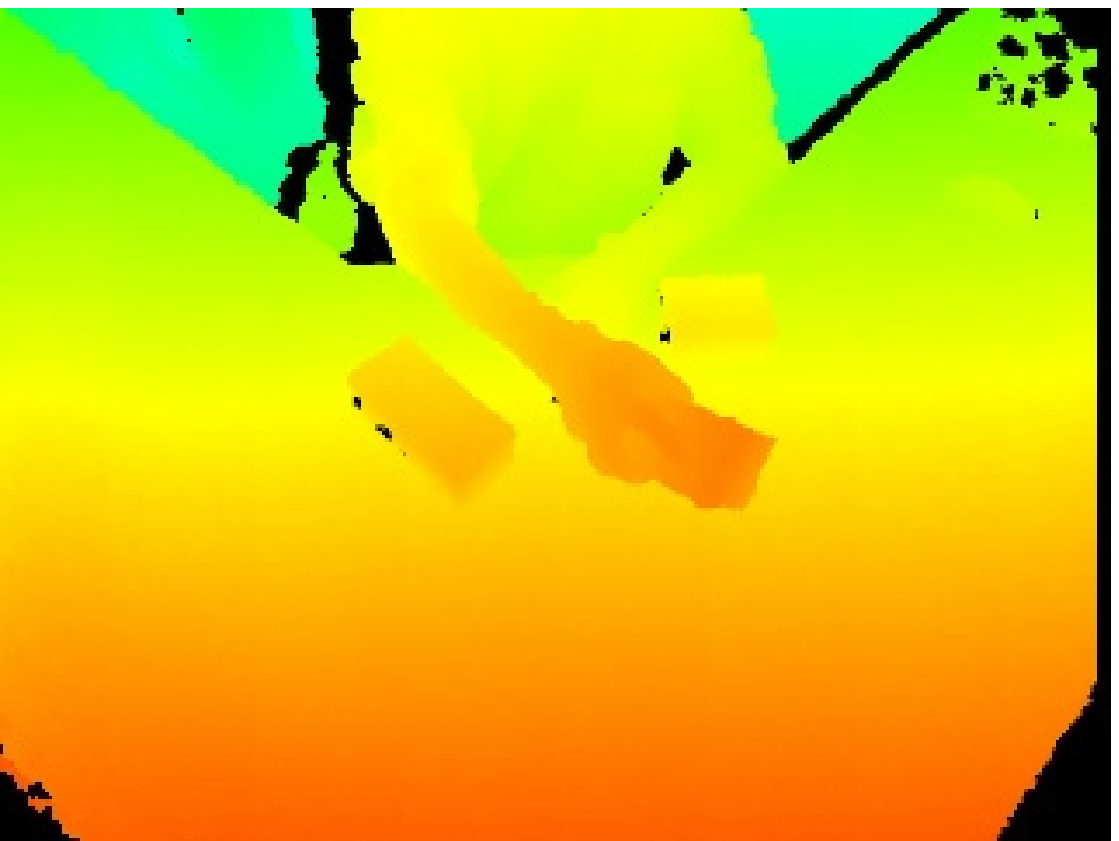, width=0.8in, height=0.5in} &
      \epsfig{file=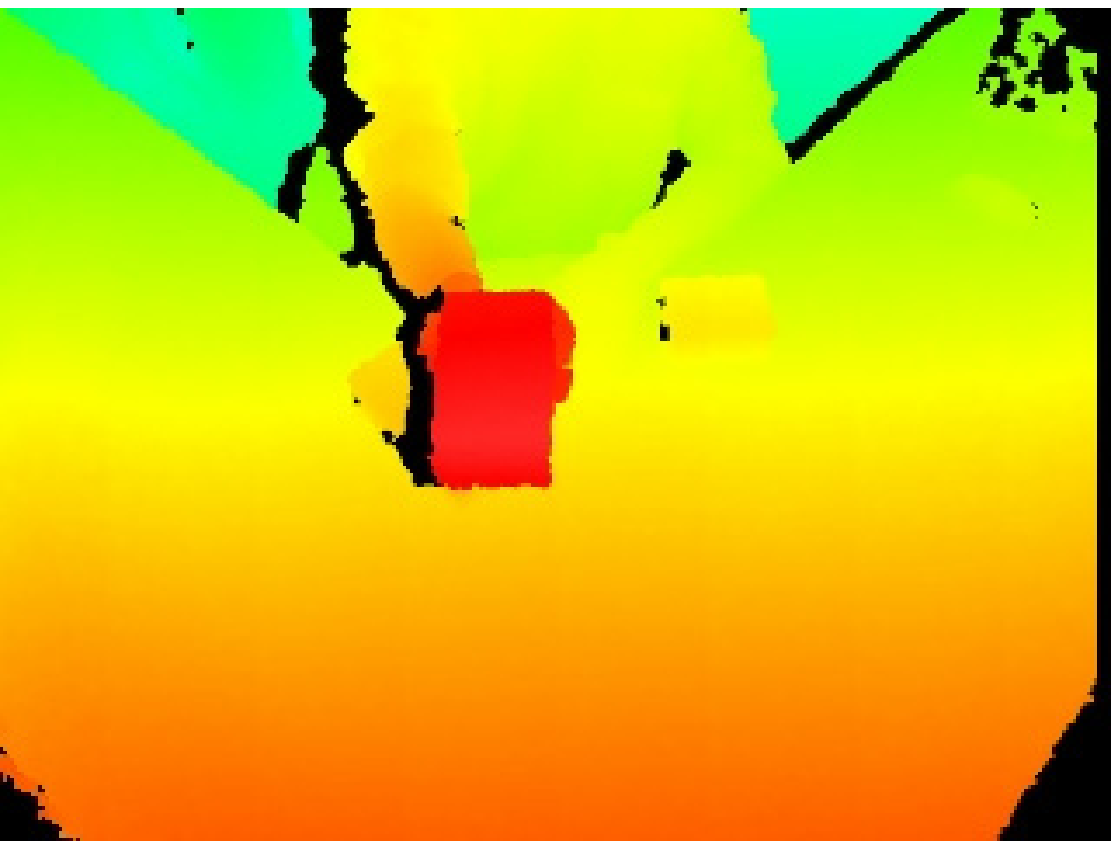, width=0.8in, height=0.5in} &
      \epsfig{file=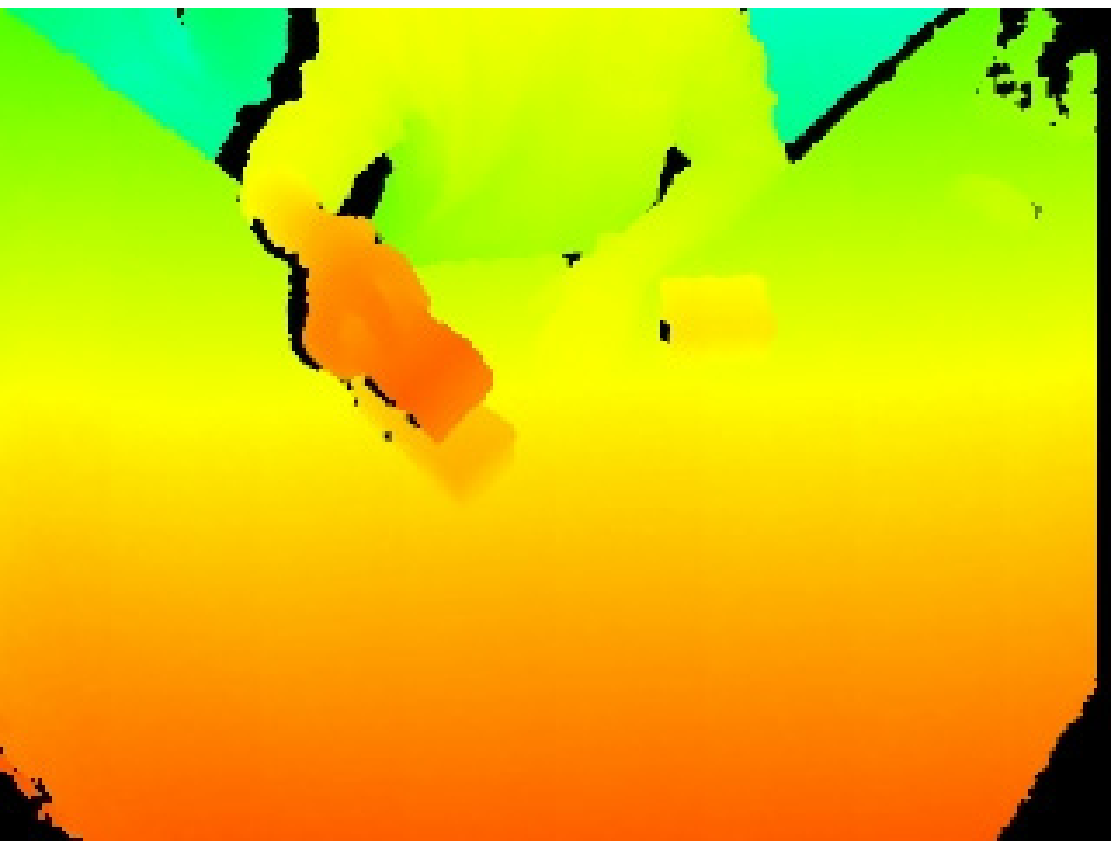, width=0.8in, height=0.5in} &
      \epsfig{file=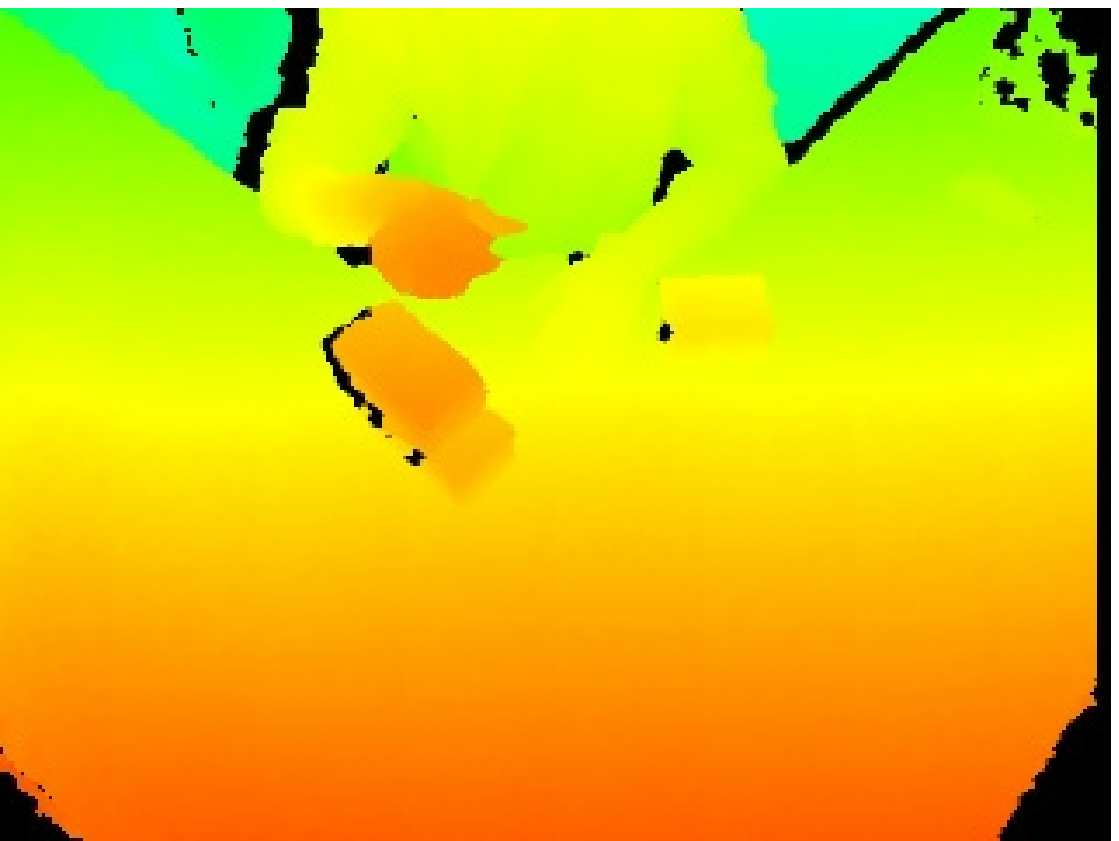, width=0.8in, height=0.5in} &
      \epsfig{file=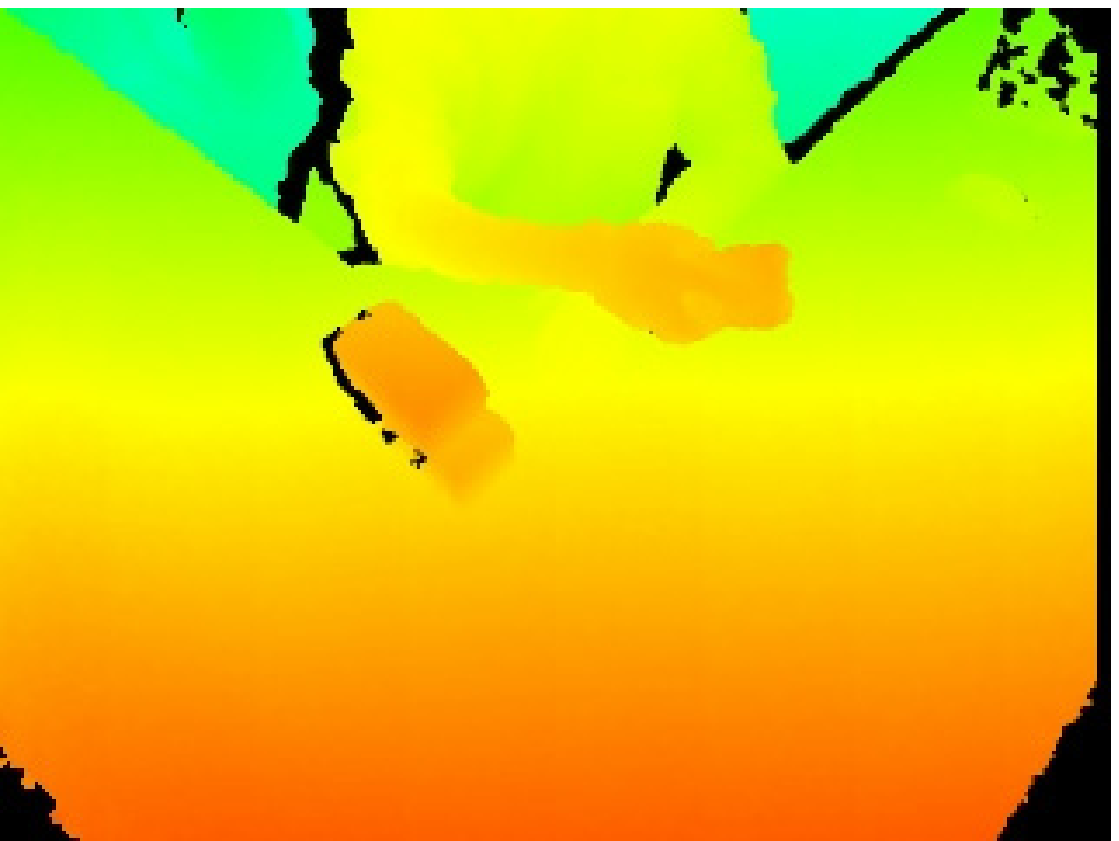, width=0.8in, height=0.5in} &
      \epsfig{file=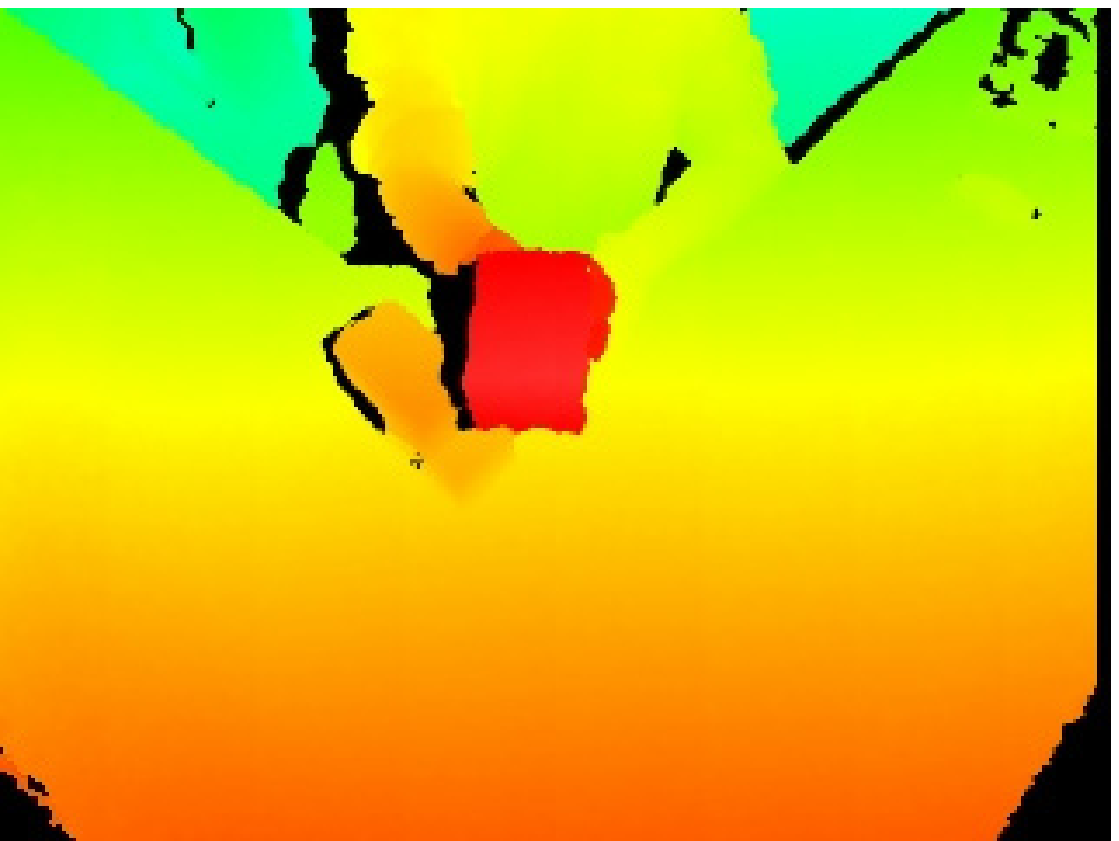, width=0.8in, height=0.5in} &
      \epsfig{file=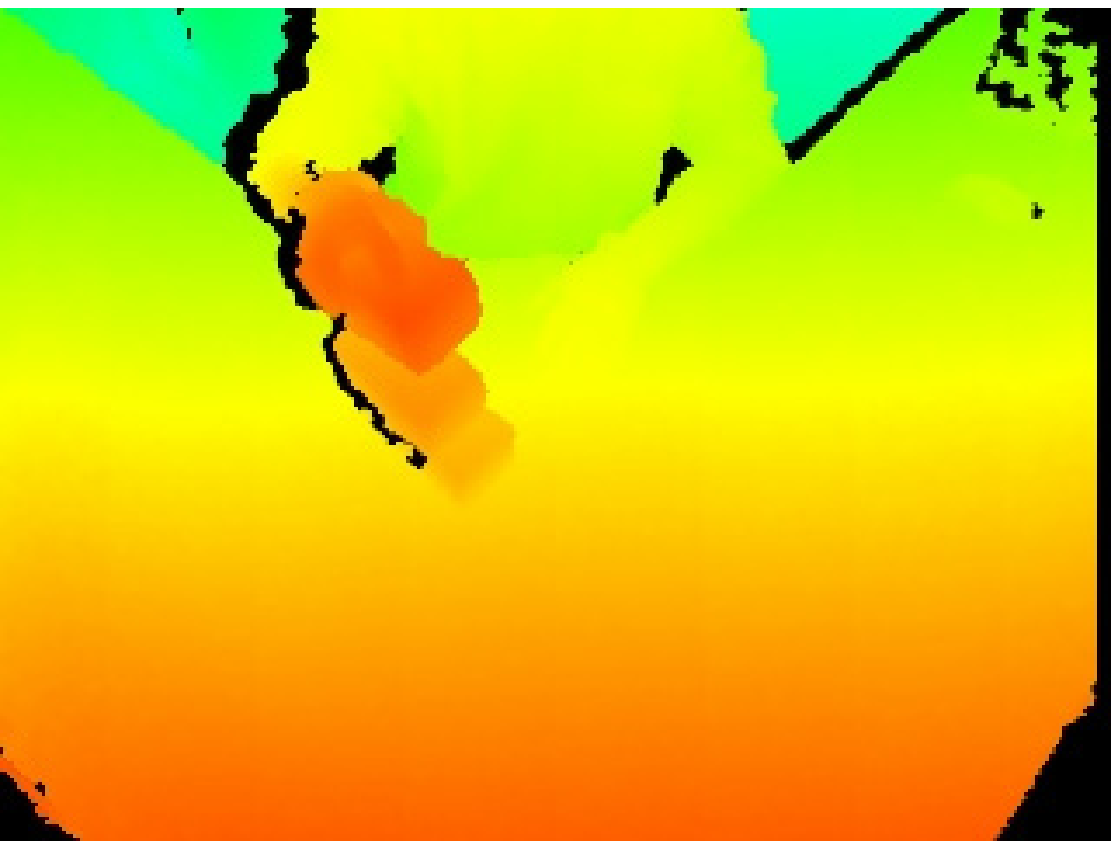, width=0.8in, height=0.5in} &
      \epsfig{file=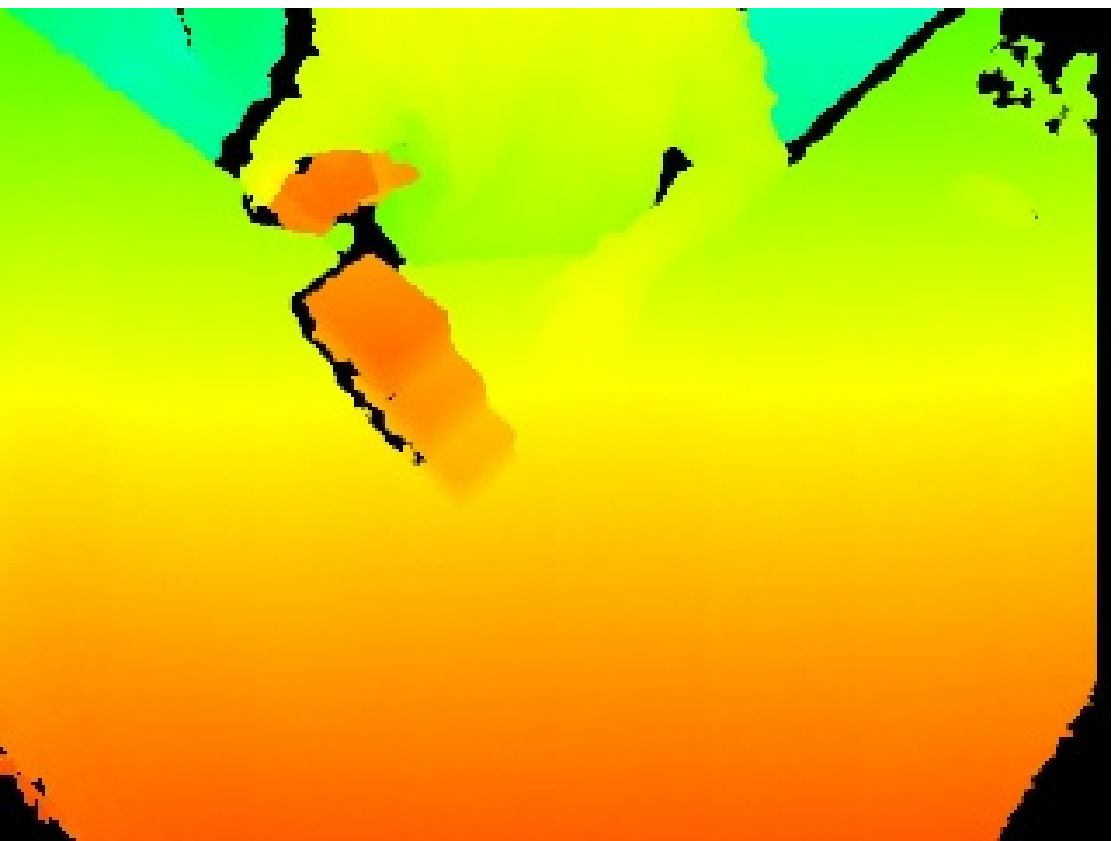, width=0.8in, height=0.5in} \\
  \end{tabular}
  \caption{Example frames from the ``stacking'' dataset. \emph{top-row: RGB images, bottom-row: aligned depth images.}}
  \label{fig_stacking}
\end{figure}

\begin{figure}[t]
  \center \small
  \addtolength{\tabcolsep}{-7pt}
  \begin{tabular}{cccccccc}
      \epsfig{file=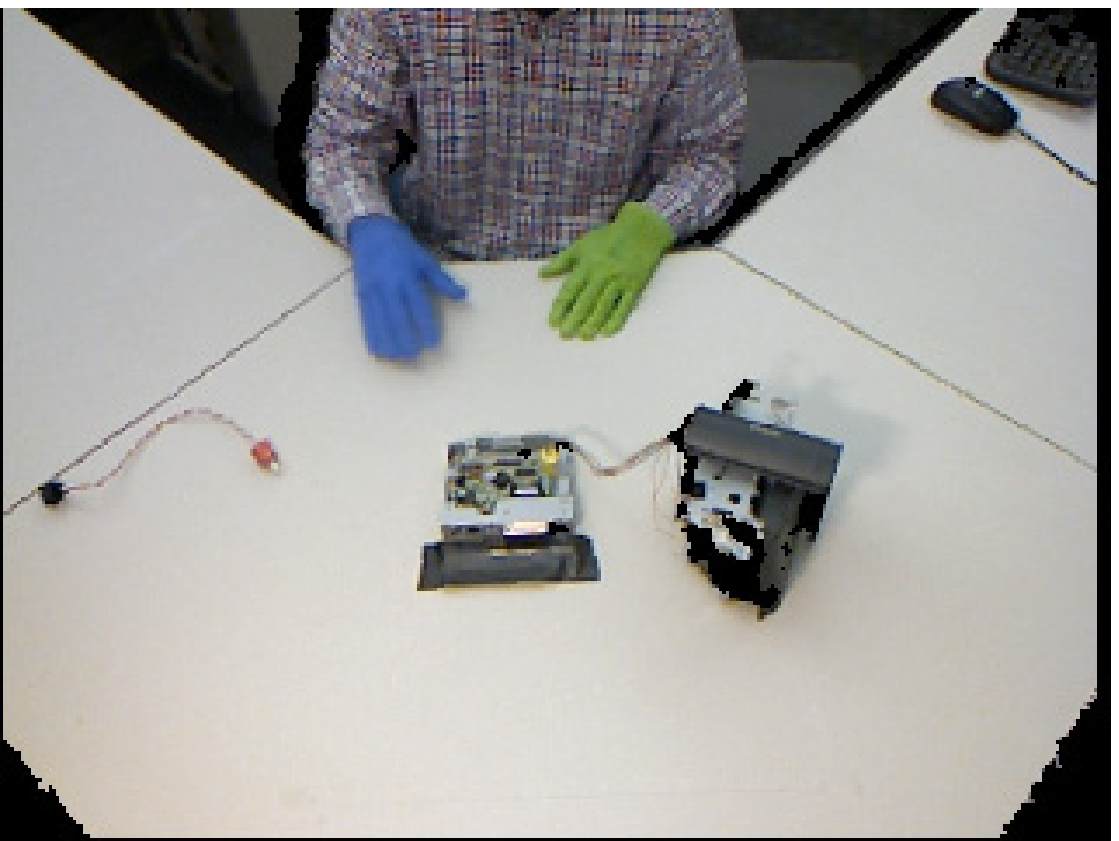, width=0.8in, height=0.5in} &
      \epsfig{file=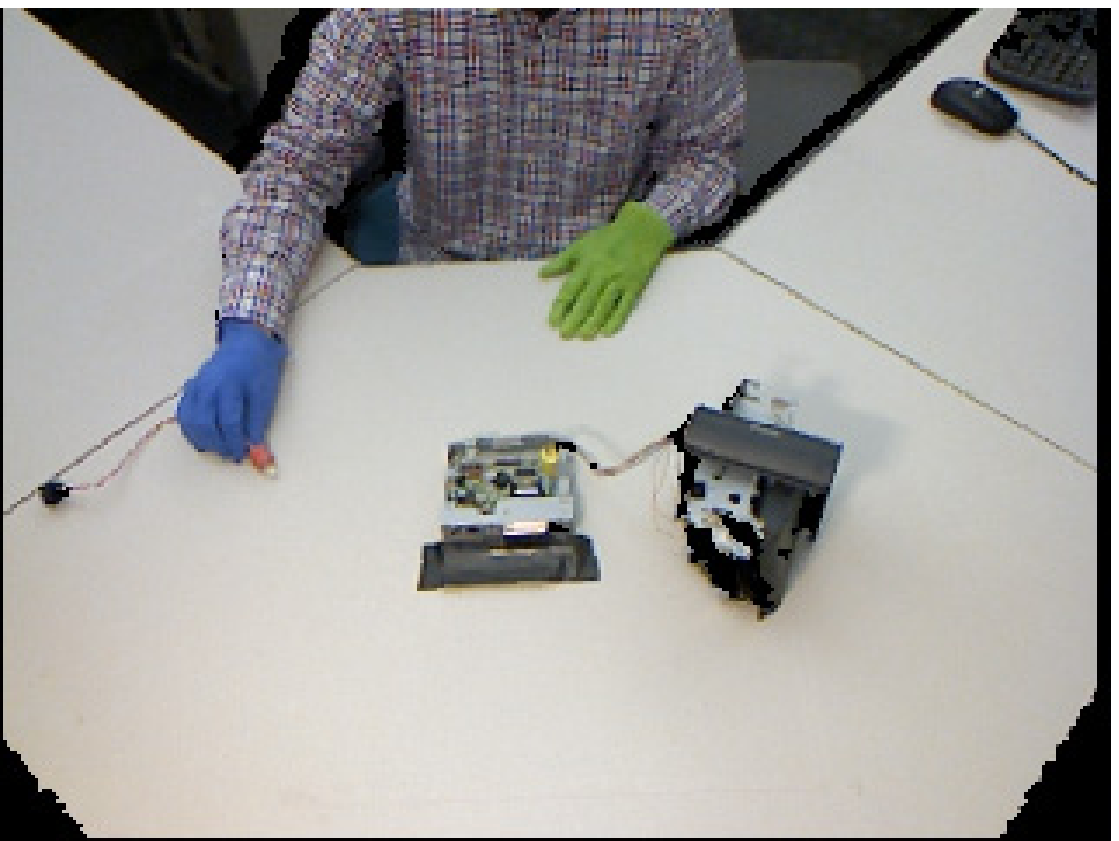, width=0.8in, height=0.5in} &
      \epsfig{file=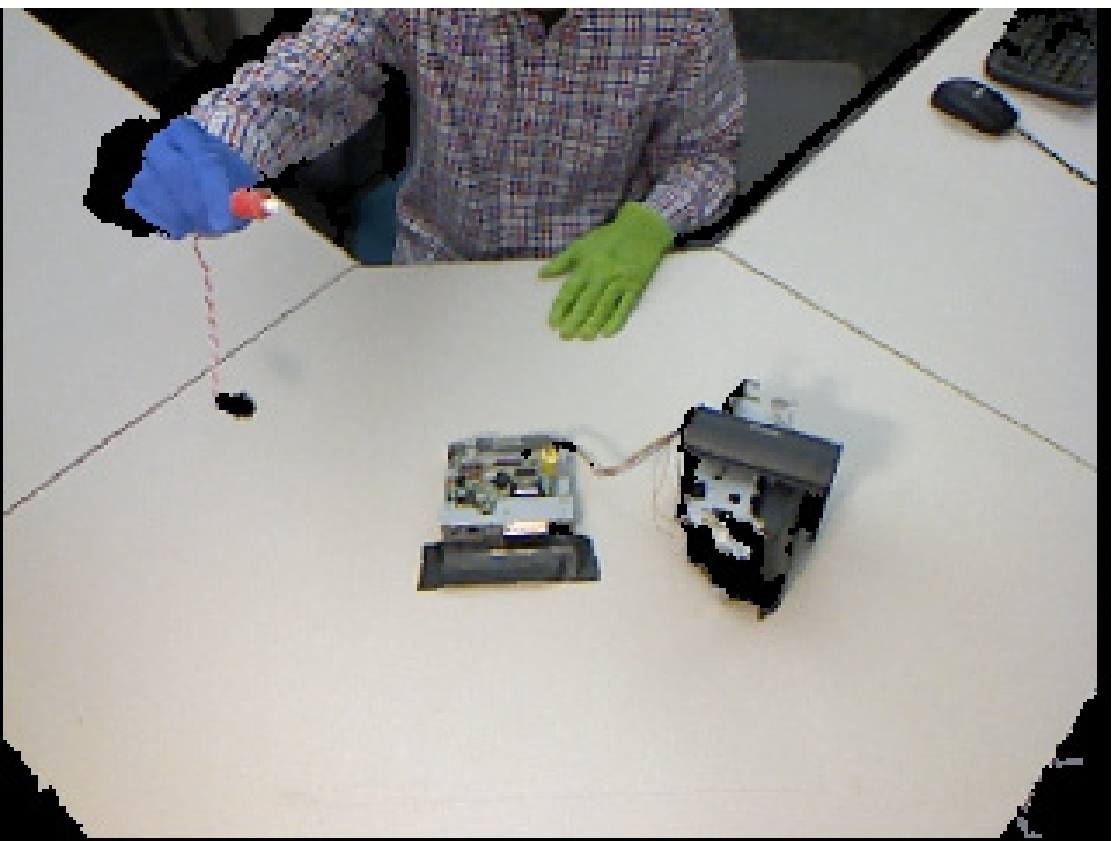, width=0.8in, height=0.5in} &
      \epsfig{file=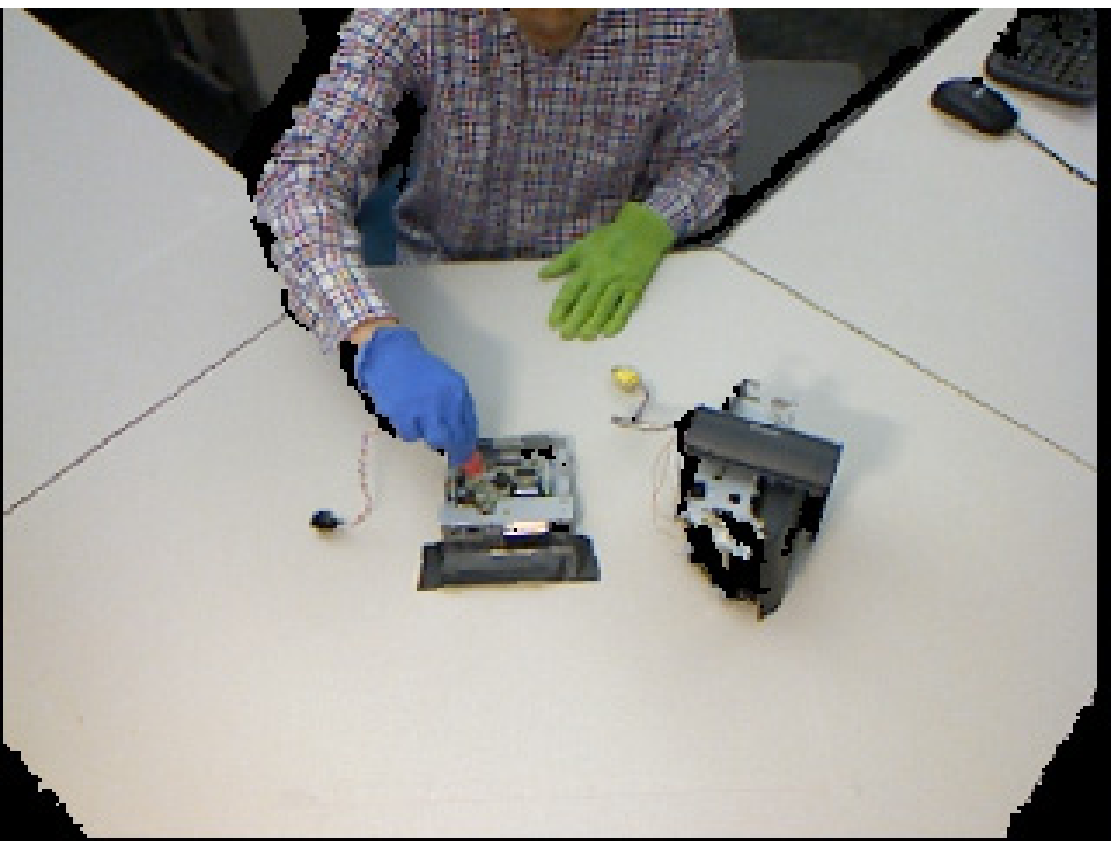, width=0.8in, height=0.5in} &
      \epsfig{file=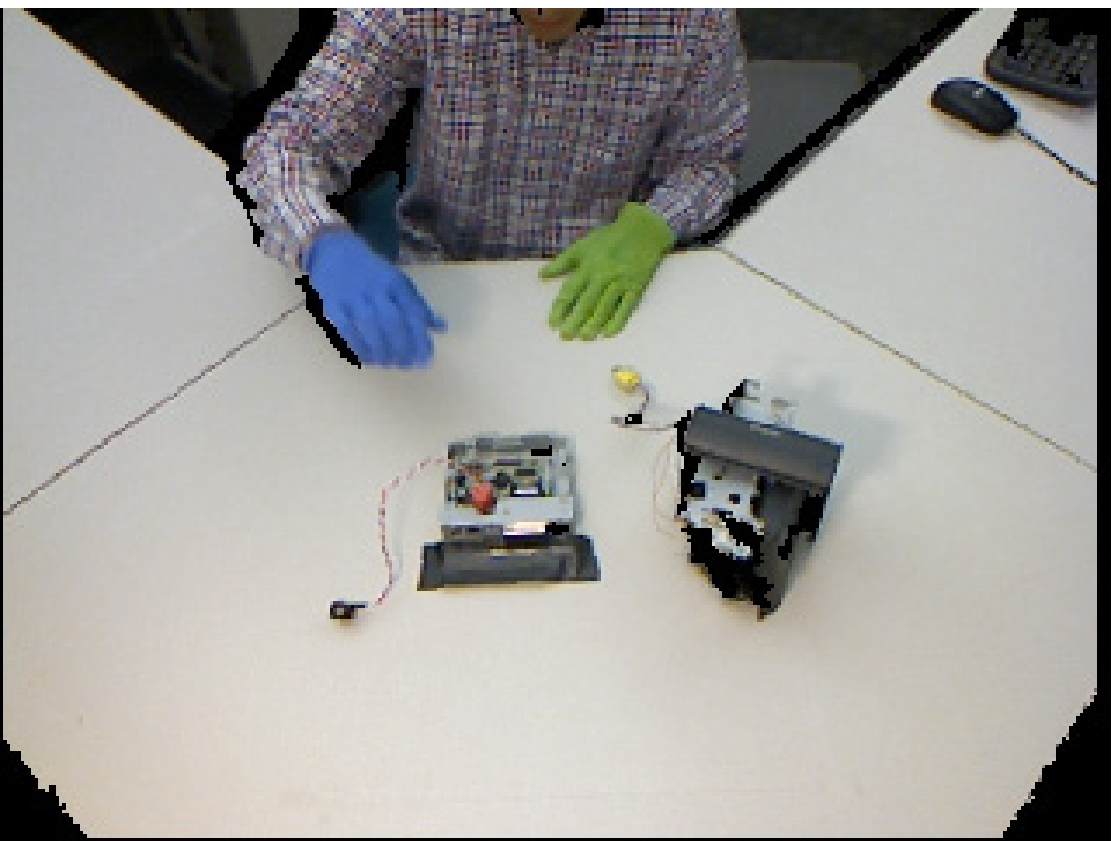, width=0.8in, height=0.5in} &
      \epsfig{file=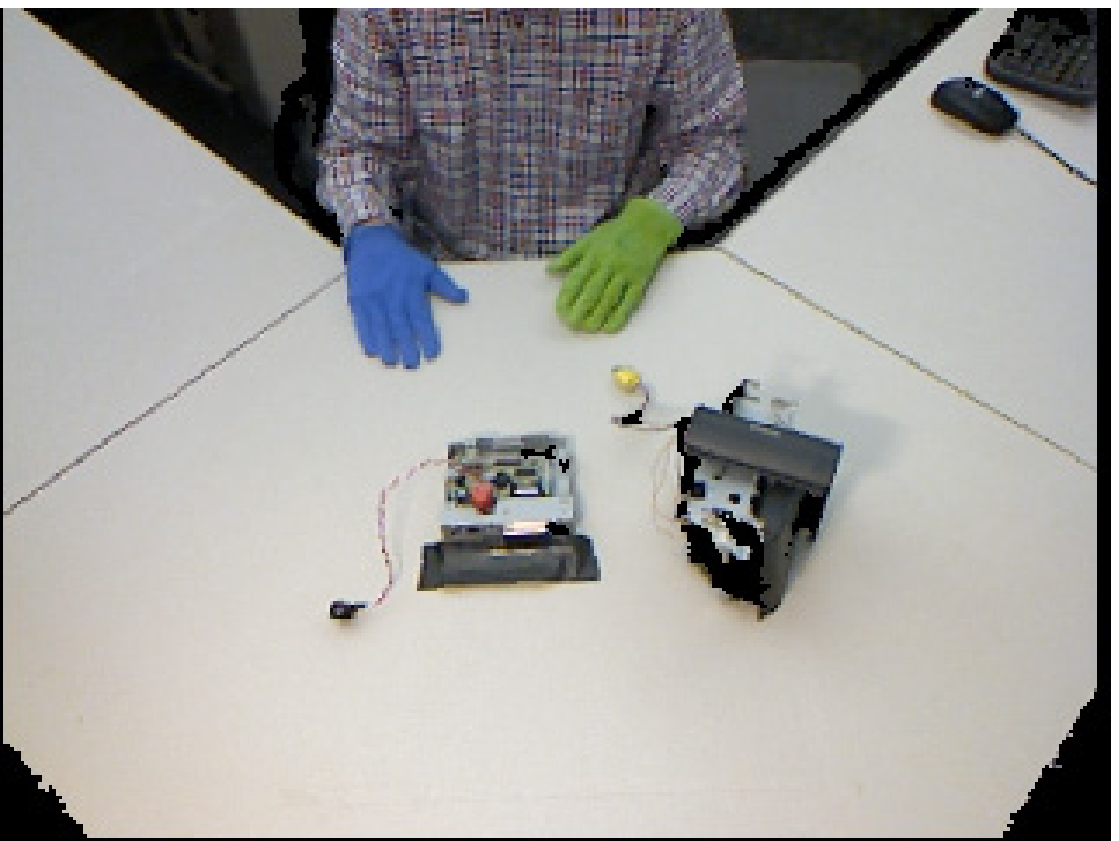, width=0.8in, height=0.5in} &
      \epsfig{file=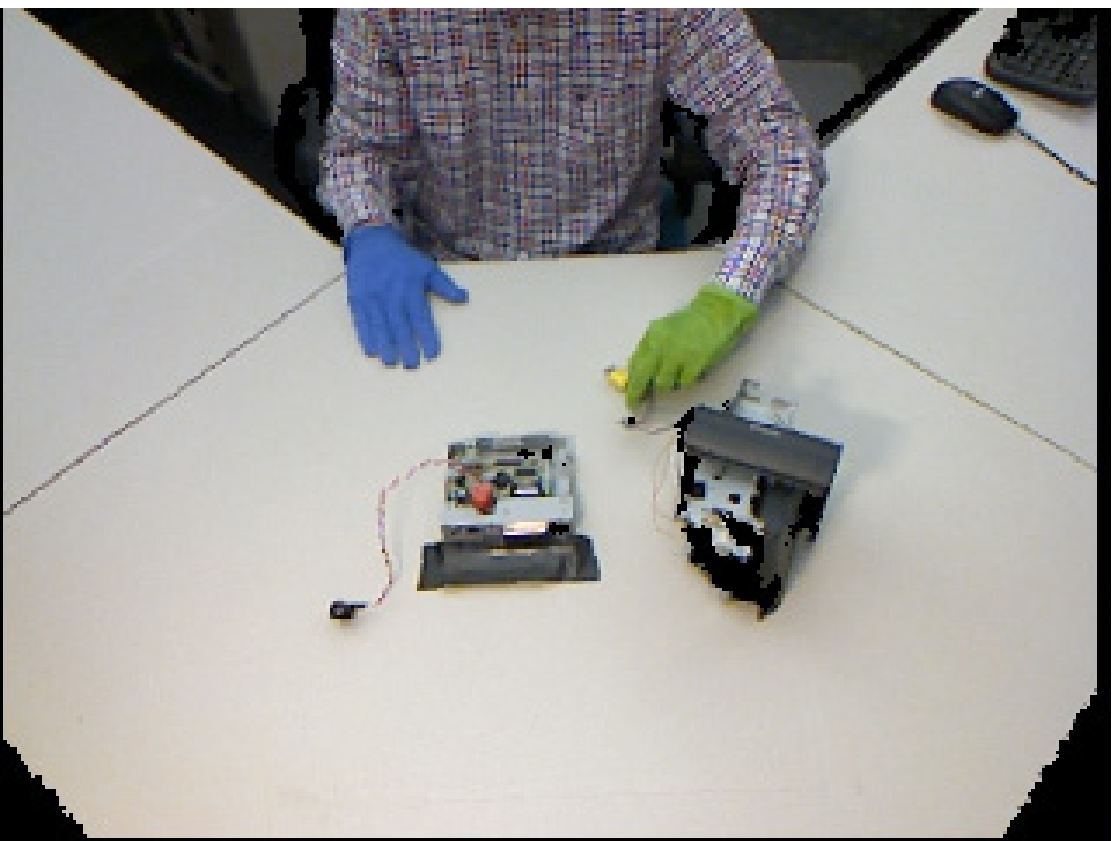, width=0.8in, height=0.5in} &
      \epsfig{file=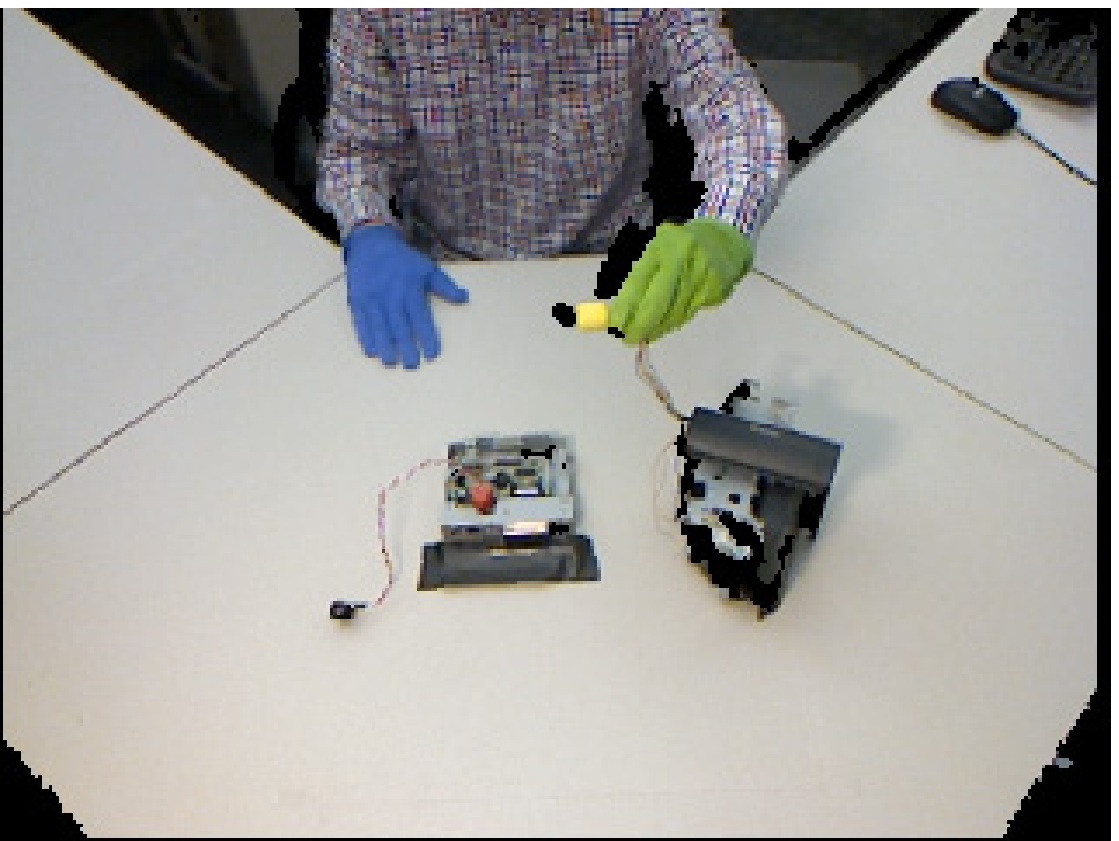, width=0.8in, height=0.5in} \\
      \epsfig{file=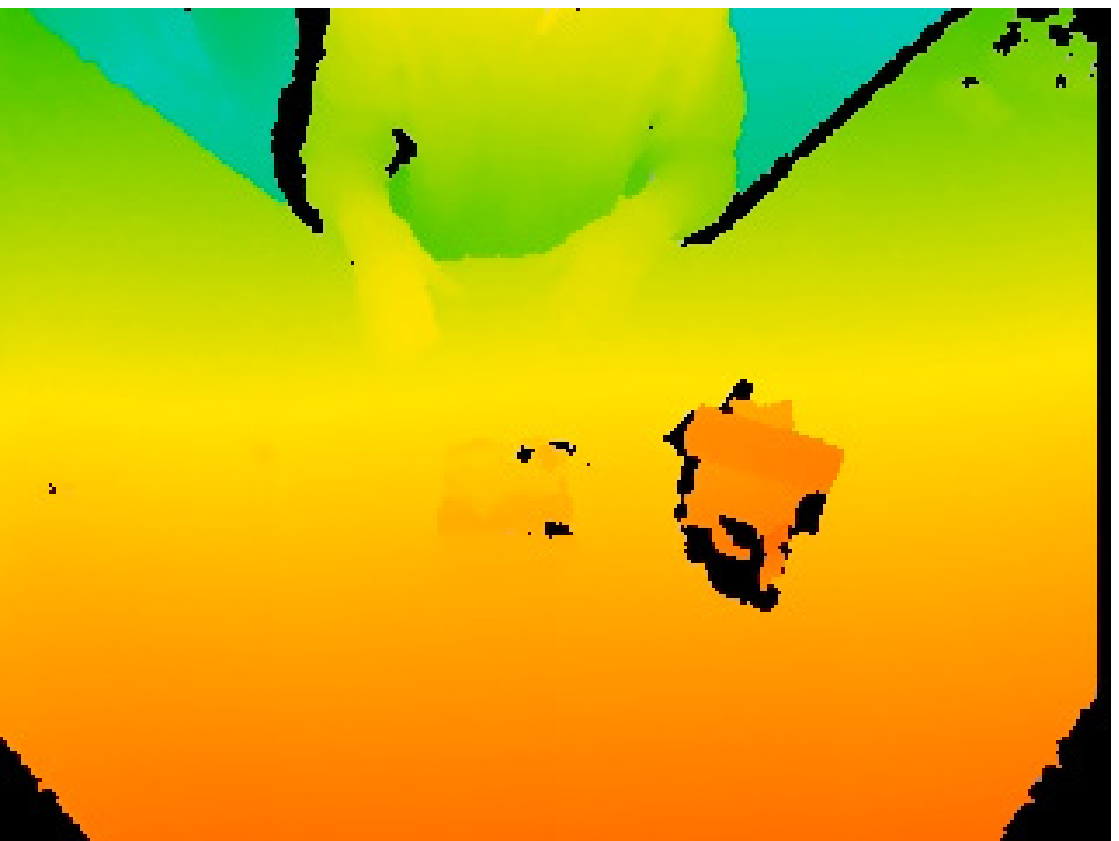, width=0.8in, height=0.5in} &
      \epsfig{file=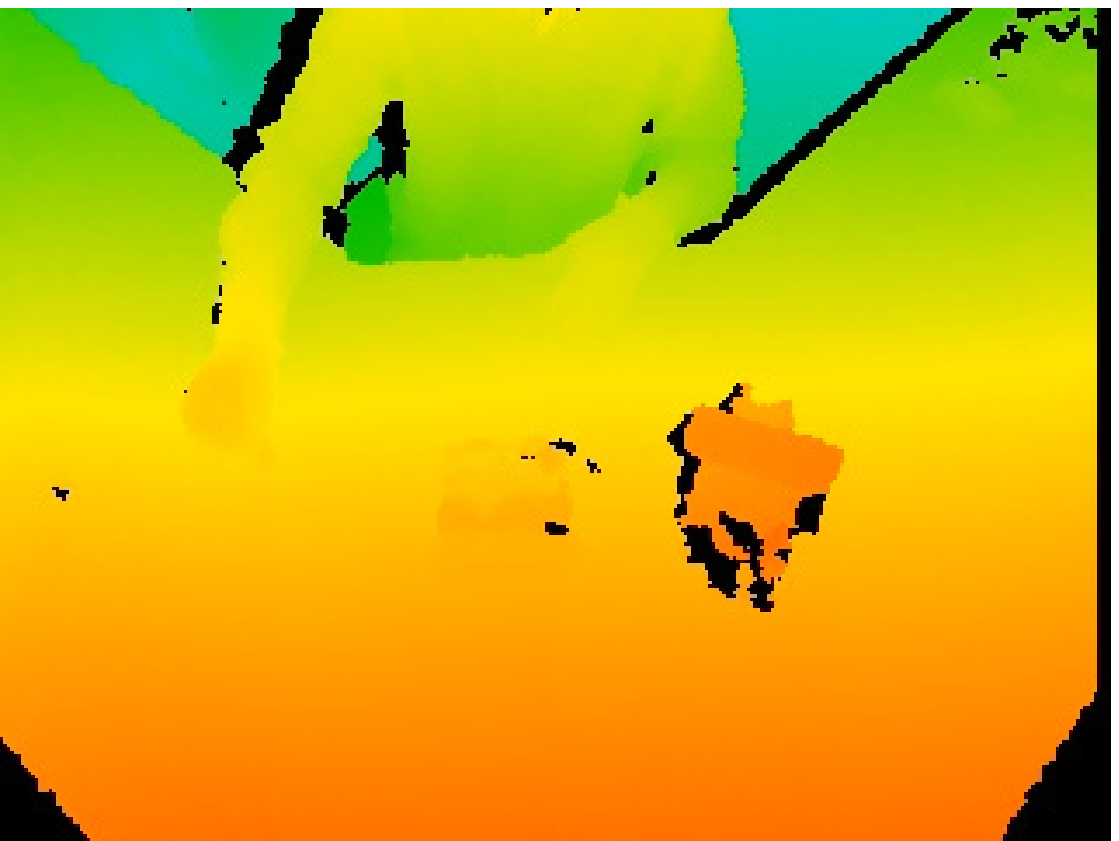, width=0.8in, height=0.5in} &
      \epsfig{file=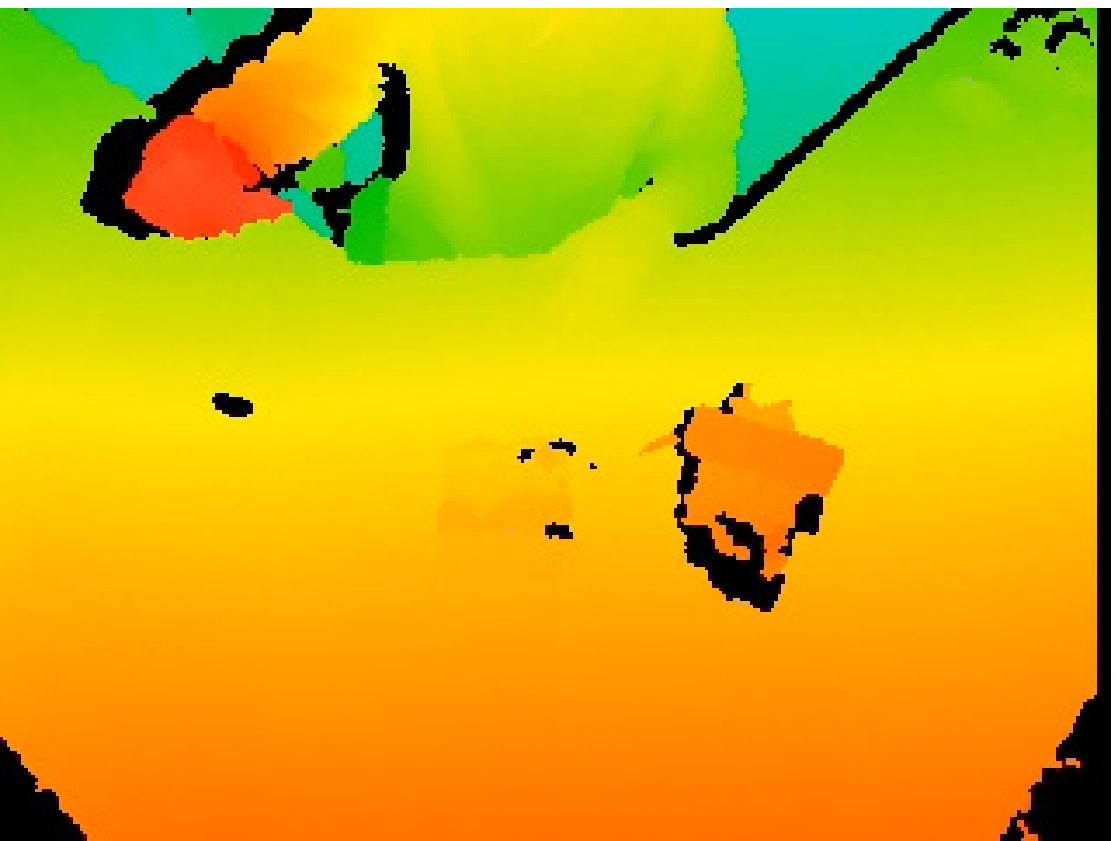, width=0.8in, height=0.5in} &
      \epsfig{file=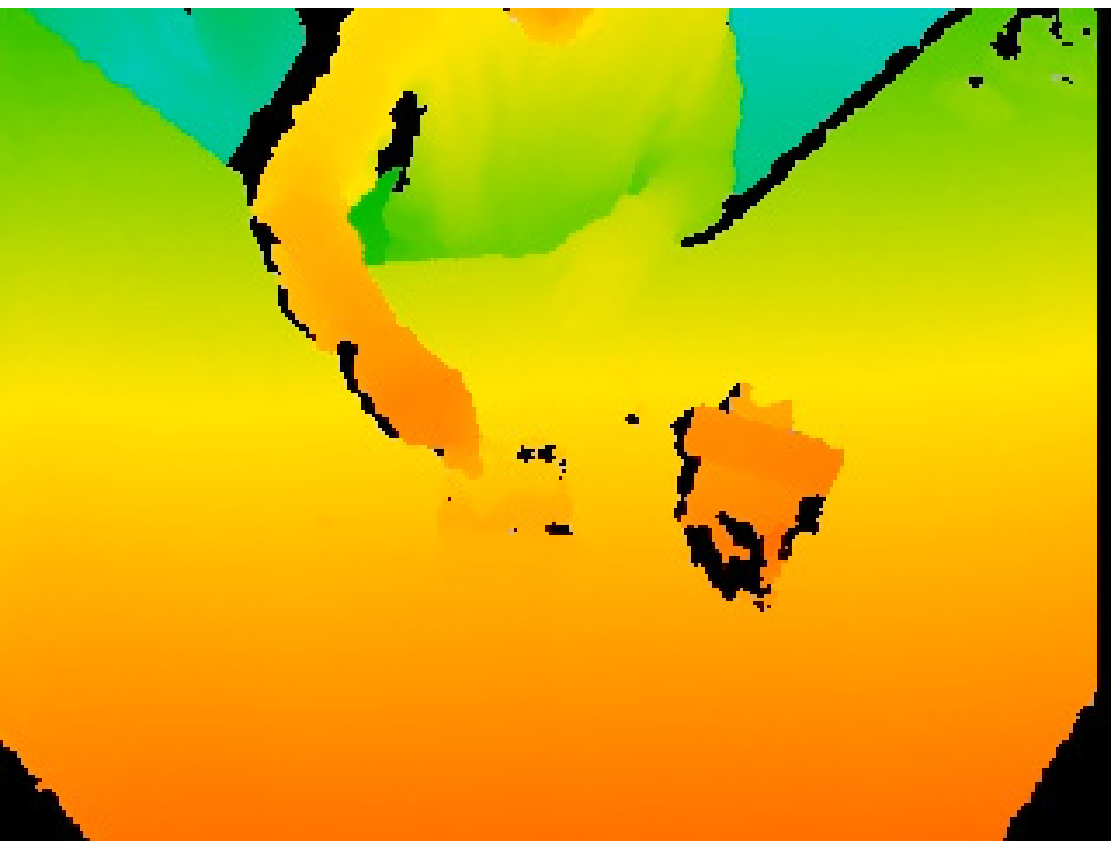, width=0.8in, height=0.5in} &
      \epsfig{file=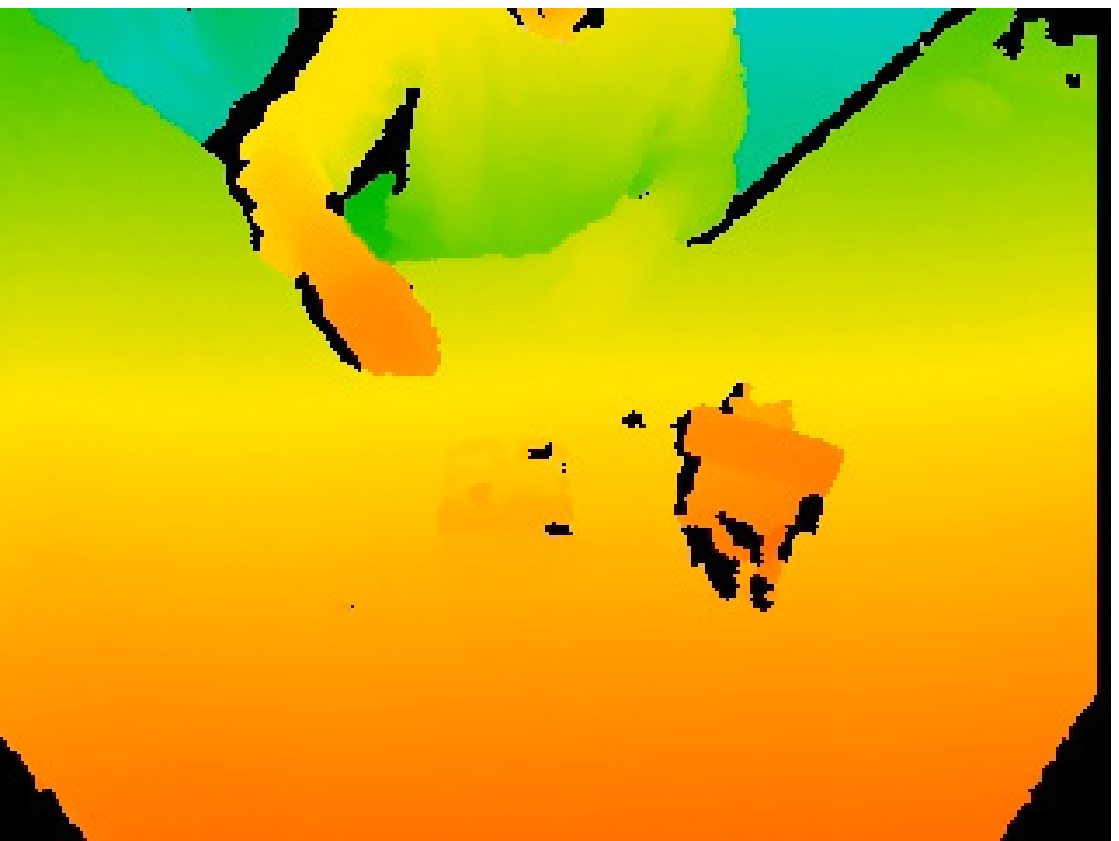, width=0.8in, height=0.5in} &
      \epsfig{file=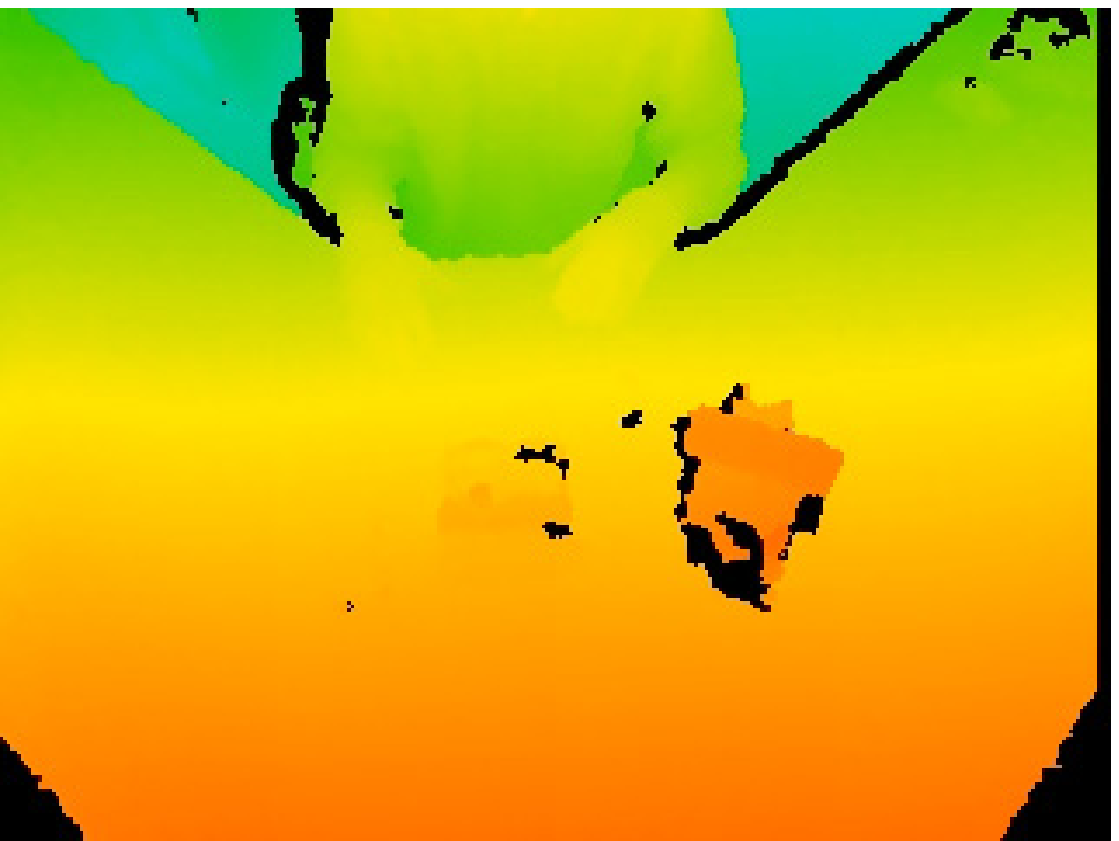, width=0.8in, height=0.5in} &
      \epsfig{file=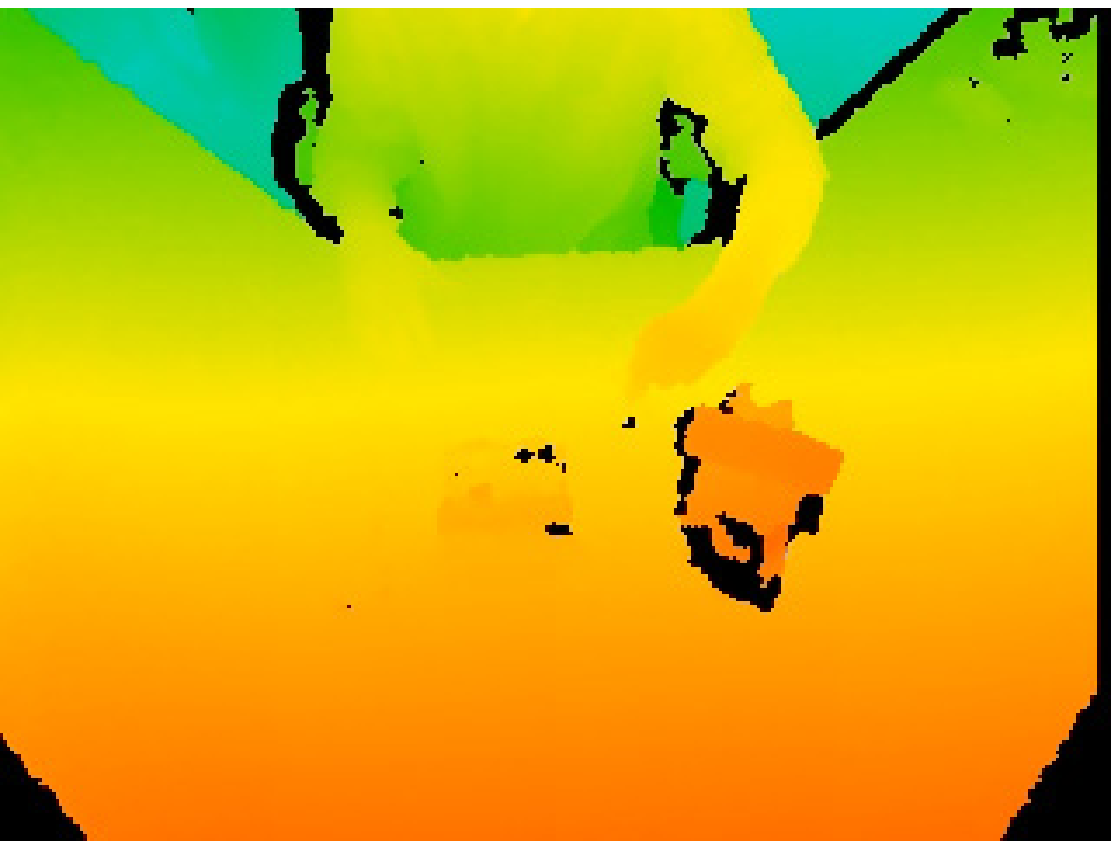, width=0.8in, height=0.5in} &
      \epsfig{file=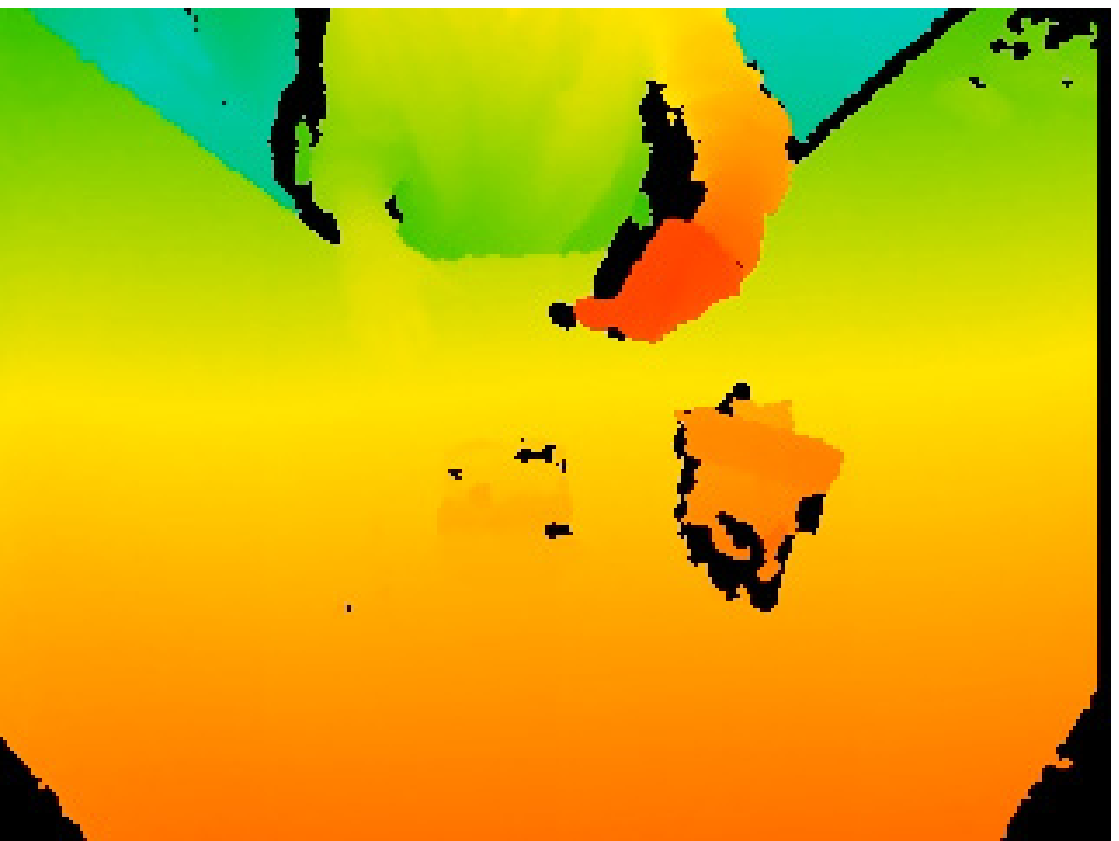, width=0.8in, height=0.5in} \\
  \end{tabular}
  \caption{Example frames from the ``assembling'' dataset. \emph{top-row: RGB images, bottom-row: aligned depth images.}}
  \label{fig_assembling}
\end{figure}

\begin{table}[t]
  \caption{Continuous action recognition for CMU MoCap dataset}\label{tab_exp_cmu}
  \centering \small
  \begin{tabular}{|c|c|c|c|c||c|c|c|}
    \hline
    SLDS & CRF & LDCRF & \cite{Ozay:Sequential}& \cite{Raptis:Spike} & STM & DBM & STM+DBM \\
    \hline
    80.0\% & 77.2\% & 82.5\% & 72.3\% & 90.9\%                      & 81.0\% & \textbf{93.3}\% & 92.1\% \\
    \hline
  \end{tabular}
  \vspace{-0.1in}
\end{table}

In addition to the above two public datasets, two in-house datasets
were also captured. The actions in these two sets feature
stronger hierarchical substructure.
The first dataset contains videos of stacking/unstacking three colored boxes,
which involves actions of ``move-arm'', ``pick-up'' and ``put-down''.
13 sequences with 567 actions were recorded in both RGB and depth videos
with one Microsoft Kinect sensor~\footnote{http://www.xbox.com/kinect}
(Fig.~\ref{fig_stacking}).
Then object tracking and 3-D reconstruction were performed to obtain the 3D trajectories of
two hands and three boxes. In this way an observation sequence in
$\mathbb{R}^{15}$ is generated. In the experiments, leave-one-out
cross-validation was performed on the 13 sequences. The continuous
recognition results are listed in Table~\ref{tab_exp_stacking}.
It is noticed that, among the four benchmark techniques, the
performance of SLDS and CRF are comparable, while LDCRF achieved the
best performance. This is reasonable because during the stacking
process, each box can be moved/stacked at any place on the desk,
which leads to large spatial variations that cannot be well modeled
by a Bayesian Network of only two layers. LDCRF applied a third
layer to capture such ``latent dynamics'', and hence achieved best
accuracy. For our proposed models, the STM alone brings LDS to a
comparable accuracy to LDCRF because it also models the substructure
transition pattern. By further incorporating duration information,
our model outperforms all other existing approaches.


The second in-house dataset is more complicated than the first one.
It involves five actions, ``move-arm'', ``pick-up'', ``put-down'',
``plug-in'' and ``plug-out'', in a printer part assembling task
(Fig.~\ref{fig_assembling}). The 3D trajectories of two hands and
two printer parts were extracted using the same Kinect sensor
system. 8 sequences were recorded and tested with leave-one-out
cross-validation. As can be seen from
Table~\ref{tab_exp_assembling}, our proposed model with both STM and
DBM outperforms other benchmark approaches by a large margin.

\begin{figure}[th]
\centering \small
 \begin{tabular}{c}
   \epsfig{file=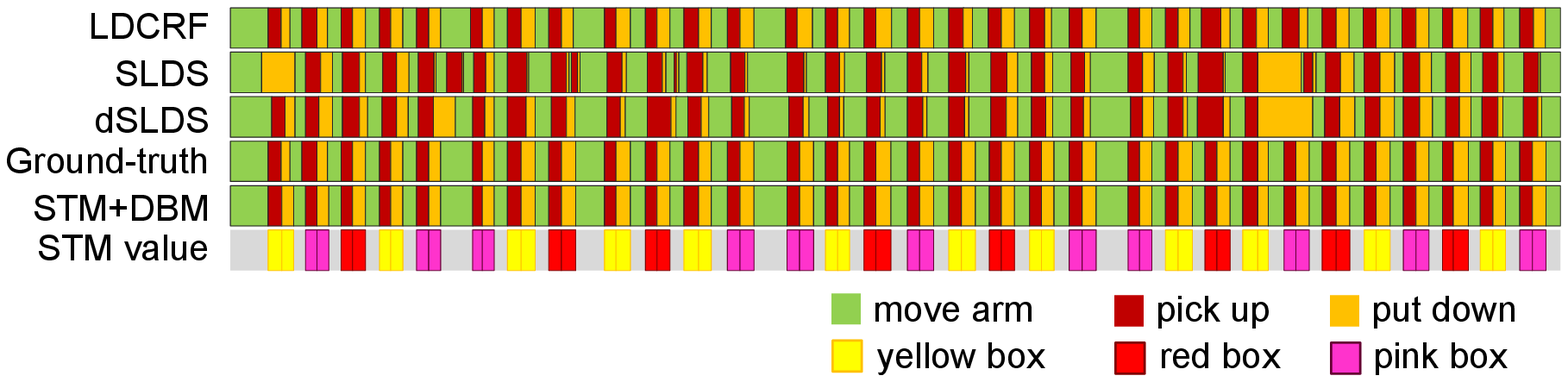, height=30mm} \\
   (a) Set I: Stacking \\
   \epsfig{file=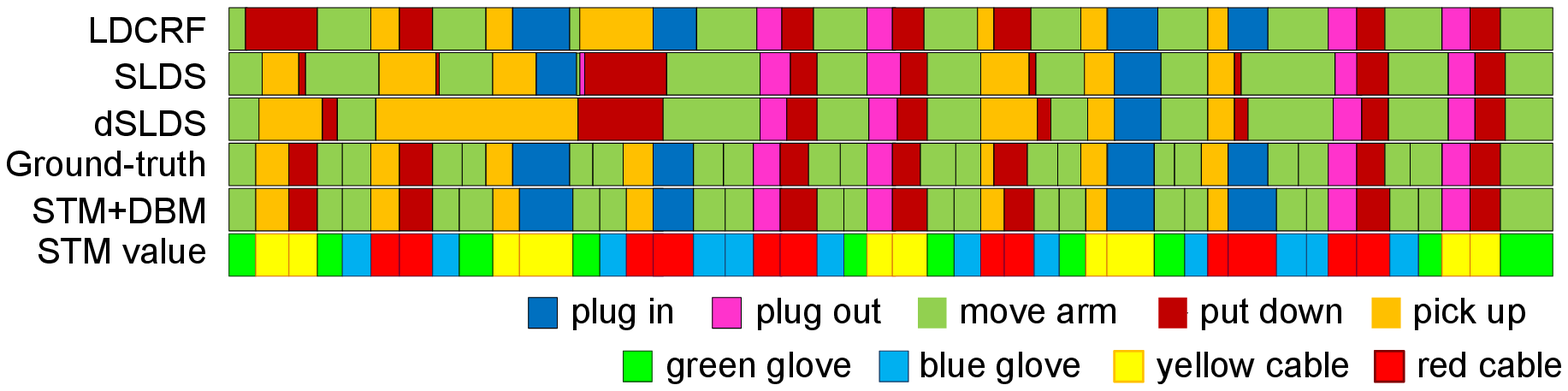, height=30mm} \\
   (b) Set II: Assembling \\
 \end{tabular}
 \caption{Continuous recognition for in-house datasets}
 \label{fig_ixmas_seq}
\end{figure}

\begin{table}[th]
  \caption{Continuous action recognition for Set I: Stacking}\label{tab_exp_stacking}
  \centering \small
  \begin{tabular}{|c|c|c||c|c|c|}
    \hline
    SLDS & CRF & LDCRF & STM & DBM & STM+DBM\\
    \hline
    64.4\% & 79.6\% & 90.3\% &88.5\% & 81.3\% & \textbf{94.4}\% \\
    \hline
  \end{tabular}
  \vspace{-0.1in}
\end{table}

\begin{table}[th]
  \caption{Continuous action recognition for Set II: Assembling}\label{tab_exp_assembling}
  \centering \small
  \begin{tabular}{|c|c|c||c|c|c|}
    \hline
     SLDS & CRF & LDCRF & STM & DBM & STM+DBM\\
    \hline
     68.2\% & 77.7\% & 88.5\% & 88.7\% & 69.0\% & \textbf{92.9}\%\\
    \hline
  \end{tabular}
  \vspace{-0.1in}
\end{table}
%

\subsection{Discussion} \label{subsec_discuss}

To provide more insightful comparison between the proposed algorithm
and other benchmark algorithms, we show two examples of continuous
action recognition results from the in-house datasets in
Fig.~\ref{fig_ixmas_seq}. The result given by SLDS contains short
and frequent switchings between incorrect action types. This is
caused by the false matching of motion patterns to an incorrect
action model. dSLDS~\cite{Sang:Parameterized} and LDCRF eliminate
the short transitions by considering additional context information;
however, their performances degrade severely around noisy or
ambiguous action periods (\eg the beginning of the sequence in
Fig.~\ref{fig_ixmas_seq}.(b)) due to false duration prior or
overdependence on discriminative classifier. Our proposed STM+DBM
approach does not suffer from any of these problems, because STM
helps to identify all action classes disregarding their variations,
and DBM further helps to improve the precision of boundaries with
both generative and discriminative duration knowledge. Another
interesting finding shown in the last rows of (a) and (b) is that
the substructure node $Z$ can be interpreted by concrete physical
meanings. For all the actions in these experiments, we find
different object involved in an action corresponds to a different
value of $Z$, which dominates the infer values $\hat{Z}_{1:T}$ in
that action. Therefore, in addition to estimating action class, we
can also find the object associated with the action by majority voting
based on $\hat{Z}_{1:T}$. In our experiments, all the inferred object
associations agree with ground truth.

\section{Conclusion and Future Work}
\label{sec:con}

In this paper, we introduce an improved SSM with two added layers
modeling the substructure transition dynamics and duration
distribution for human action.
The first layer encodes the sparse and global temporal transition
structure of action primitives and also maintains action variations.
The second layer injects discriminative information into a logistic duration model
and discovers action boundaries more adaptively.
We design a Rao-Blackwellised particle filter for efficient inference.
Our comprehensive experimental results validate the effectiveness of both two
layers of our model in continuous action recognition.
As future work we plan to apply our model to actions in less constrained
scenarios and use more advanced low-level descriptors to deal with
unreliable observations.

\appendix

\section{Derivation of Eq.\eqref{eq:objfunc_blktrans}}
\label{sec:app_blktrans}

With the constraint of Eq.\eqref{eq:constraint_g}, we have
\begin{align}
    &   p(Z_{t+1}^j|Z_t^i) \nonumber \\
    = & \sum \limits_{r, q} p(Q_{t+1}^r, Z_{t+1}^j | Q_t^q, Z_t^i) p(Q_t^q | Z_t^i) \nonumber \\
    = & p(Q_{t+1}^{g(j)}, Z_{t+1}^j | Q_t^{g(i)}, Z_t^i) \nonumber \\
        \label{eq:tran_cnvrt}
    = & p(Q_{t+1}^{g(j)} | Q_t^{g(i)}) p(Z_{t+1}^j | Q_{t+1}^{g(j)}, Z_t^i) \\
    = & p(Q_{t+1}^{g(j)} | Q_t^{g(i)}) \frac{p(Z_{t+1}^j, Q_{t+1}^{g(j)} | Z_t^i)}{p(Q_{t+1}^{g(j)} | Z_t^i)}\nonumber \\
    = & p(Q_{t+1}^{g(j)} | Q_t^{g(i)}) \frac{p(Q_{t+1}^{g(j)} | Z_t^i, Z_{t+1}^j) p(Z_{t+1}^j | Z_t^i)}
                                  {\sum_{j'} p(Q_{t+1}^{g(j)} | Z_t^i, Z_{t+1}^{j'}) p(Z_{t+1}^{j'} | Z_t^i)} \nonumber \\
    = & p(Q_{t+1}^{g(j)} | Q_t^{g(i)}) \frac{p(Z_{t+1}^j | Z_t^i)}
                                   {\sum_{j' \in \mathcal{G}(g(j))} p(Z_{t+1}^{j'} | Z_t^i)} \nonumber
\end{align}
Equivalently,
\begin{equation}
\label{eq:cond_QZ2Z}
    p(Q_{t+1}^{g(j)} | Q_t^{g(i)}) = \sum_{j' \in \mathcal{G}(g(j))} p(Z_{t+1}^{j'} | Z_t^i), \;\;\; \forall i, j
\end{equation}
Eq.\eqref{eq:tran_cnvrt} shows that we can eliminate $Q$ from the substructure transition model,
which results in the simplified objective function in Eq.\eqref{eq:objfunc_blktrans}.
Eq.\eqref{eq:cond_QZ2Z} leads to the equality constraint in Eq.\eqref{eq:objfunc_blktrans}.

\section{Derivation of Eq.\eqref{eq:theta_hat}}
\label{sec:app_thetahat}

From the KKT conditions, we have:
\begin{eqnarray}
  &             & \xi_{ij} - (\lambda_{i, g(j)} - \gamma_i + \mu_{ij})\hat{\theta}_{ij} = 0 \nonumber \\
  & \Rightarrow & \xi_{ij} - (\lambda_{i, g(j)} - \gamma_i)\hat{\theta}_{ij} = 0 \nonumber \\
  & \Rightarrow & \hat{\theta}_{ij} \propto \xi_{ij}, \;\;\; \forall i, r, j \in \mathcal{G}(r) \nonumber \\
  & \Rightarrow & \hat{\theta}_{ij} = \hat{\phi}_{g(i), g(j)} \frac{\xi_{ij}}{\sum_{j' \in \mathcal{G}(g(j))} \xi_{ij'}} \nonumber
\end{eqnarray}
\begin{eqnarray}
  &             & -\alpha_{qr} + \Sigma_{i \in \mathcal{G}(q)} \lambda_{ir}\hat{\phi}_{qr} = 0 \nonumber \\
  & \Rightarrow & -\alpha_{qr} + \Sigma_{i \in \mathcal{G}(q), j \in \mathcal{G}(r)} \lambda_{ir} \hat{\theta}_{ij} = 0 \nonumber \\
  & \Rightarrow & -\alpha_{qr} + \Sigma_{i \in \mathcal{G}(q), j \in \mathcal{G}(r)}
                  \left( \xi_{ij} + \gamma_i \hat{\theta}_{ij} \right) = 0 \nonumber \\
  & \Rightarrow & \Sigma_{i \in \mathcal{G}(q), j \in \mathcal{G}(r)} \xi_{ij} -\alpha_{qr} +
                  \Sigma_{i \in \mathcal{G}(q)} \gamma_i \hat{\phi}_{qr} = 0 \nonumber \\
  & \Rightarrow & \hat{\phi}_{qr} \propto \Sigma_{i \in \mathcal{G}(q), j \in \mathcal{G}(r)} \xi_{ij} -\alpha_{qr} \nonumber
\end{eqnarray}
Note that $\hat{\phi}_{qr} \geq 0, \sum_{r}\hat{\phi}_{qr}=1$, and we obtain the second equation in
Eq.\eqref{eq:theta_hat}.

\section{Derivation of Eq.\eqref{eq:rbpf_sdz} and \eqref{eq:int_norm_logit}}
\label{sec:app_rbpfsdz}

Denote $\mathbf{X}_t=(S_t, D_t, Z_t, X_t)$, and we have:
\begin{align}
      & p(\mathbf{X}_t | \mathbf{y}_{1:t}) \nonumber \\
    = & \int \frac{p(\mathbf{y}_t|\mathbf{X}_t) p(\mathbf{X}_t|\mathbf{X}_{t-1}) p(\mathbf{X}_{t-1} | \mathbf{y}_{1:t-1})}
                  {p(\mathbf{y}_t|\mathbf{y}_{1:t-1})} d \mathbf{X}_{t-1} \nonumber \\
    \propto & \int \sum_n  w^{(n)}_{t-1} \delta_{S_{t-1}}(s^{(n)}_{t-1}) \delta_{D_{t-1}}(d^{(n)}_{t-1})
            \delta_{Z_{t-1}}(z^{(n)}_{t-1}) \nonumber \\
            & \;\; \times \chi_{t-1}^{(n)}(X_{t-1}) p(\mathbf{y}_t|\mathbf{X}_t) p(\mathbf{X}_t|\mathbf{X}_{t-1})
            d \mathbf{X}_{t-1} \nonumber \\
    \propto & \sum_n  w^{(n)}_{t-1} p(\mathbf{y}_t|S_t, Z_t, X_t) p(S_t|D_t, s^{(n)}_{t-1}) \nonumber \\
            & \;\; \times p(Z_t | S_t, D_t, z^{(n)}_{t-1}) \int p(D_t | d^{(n)}_{t-1}, s^{(n)}_{t-1}, z^{(n)}_{t-1}, \mathbf{x}_{t-1}) \nonumber \\
            & \;\; \times p(X_t|\mathbf{x}_{t-1}, S_t, Z_t) \chi_{t-1}^{(n)}(\mathbf{x}_{t-1}) d \mathbf{x}_{t-1} \nonumber
\end{align}
Taking integral with respect to $X_t$, we get:
\begin{align}
            & p(S_t, D_t, Z_t | \mathbf{y}_{1:t}) \nonumber \\
    \propto & \sum_n w^{(n)}_{t-1} p(S_t| D_t, s^{(n)}_{t-1}) p(Z_t | S_t, D_t, z^{(n)}_{t-1}) \nonumber \\
            & \;\; \times \int \int p(\mathbf{y}_t|S_t, Z_t, \mathbf{x}_t) p(\mathbf{x}_t|\mathbf{x}_{t-1}, S_t, Z_t) d \mathbf{x}_t \nonumber \\
            & \;\; \times p(D_t|d^{(n)}_{t-1}, s^{(n)}_{t-1}, z^{(n)}_{t-1}, \mathbf{x}_{t-1}) \chi_{t-1}^{(n)}(\mathbf{x}_{t-1})
                d \mathbf{x}_{t-1} \nonumber
\end{align}
Eq.\eqref{eq:rbpf_sdz} and \eqref{eq:int_norm_logit} can be obtained by
replacing the inner integral with $p(\mathbf{y}_t|\mathbf{x}_{t-1}, S_t, Z_t)$.

\section{Evaluation of Eq.\eqref{eq:int_norm_logit}}
\label{sec:app_logit}

From Eq.\eqref{eq:lds_y} and Eq.\eqref{eq:lds_x}, we have:
\begin{align}
     p(\mathbf{y}_t | S^i_t, Z^j_t, \mathbf{x}_t)   & = \mathcal{N}(\mathbf{y}_t; \mathbf{B}^{ij} \mathbf{x}_t, \mathbf{R}^{ij}) \nonumber \\
     p(\mathbf{x}_t|\mathbf{x}_{t-1}, S^i_t, Z^j_t) & = \mathcal{N}(\mathbf{x}_t; \mathbf{A}^{ij} \mathbf{x}_{t-1},  \mathbf{Q}^{ij}) \nonumber
\end{align}
which leads to:
\begin{align}
      & p(\mathbf{y}_t | \mathbf{x}_{t-1}, S^i_t, Z^j_t) \nonumber \\
    = & \mathcal{N}(\mathbf{y}_t; \mathbf{B}^{ij} \mathbf{A}^{ij} \mathbf{x}_{t-1},
                    \mathbf{B}^{ij} \mathbf{Q}^{ij} {\mathbf{B}^{ij}}^T + \mathbf{R}^{ij}) \nonumber \\
    = & \mathcal{N}(\mathbf{y}_t; \bm{\mu}_Y, \mathbf{\Sigma}_Y)
       =\mathcal{N}(\mathbf{y}_t; \mathbf{A}\mathbf{x}_{t-1}, \mathbf{\Sigma}_Y)   \nonumber
\end{align}
where $\bm{\mu}_Y \triangleq \mathbf{B}^{ij} \mathbf{A}^{ij} \mathbf{x}_{t-1}$,
$\mathbf{\Sigma}_Y \triangleq \mathbf{B}^{ij} \mathbf{Q}^{ij} {\mathbf{B}^{ij}}^T + \mathbf{R}^{ij}$,
and $\mathbf{A} \triangleq \mathbf{B}^{ij} \mathbf{A}^{ij}$.
We also define, $\bm{\mu}_X \triangleq \hat{\mathbf{x}}^{(n)}_{t-1}$,
$\mathbf{\Sigma}_X \triangleq \mathbf{P}^{(n)}_{t-1}$, and have:
\begin{equation}
    \chi_{t-1}^{(n)}(\mathbf{x}_{t-1}) = \mathcal{N}(\mathbf{x}_{t-1}; \bm{\mu}_X, \mathbf{\Sigma}_X) \nonumber
\end{equation}

The above two Gaussian distributions, \ie the first two terms in the integral of Eq.\eqref{eq:int_norm_logit},
can be combined as a single Gaussian of $\mathbf{x}_{t-1}$.
Omit all the subscriptions, and the product of exponential terms is:
\begin{align}
      & (\mathbf{x}-\bm{\mu}_X)^T \mathbf{\Sigma}^{-1}_X (\mathbf{x}-\bm{\mu}_X)
       + (\mathbf{y}-\mathbf{A}\mathbf{x})^T \mathbf{\Sigma}^{-1}_Y (\mathbf{y}-\mathbf{A}\mathbf{x}) \nonumber \\
    = & \mathbf{x}^T\mathbf{\Sigma}^{-1}_X\mathbf{x} + \mathbf{x}^T\mathbf{A}^T\mathbf{\Sigma}^{-1}_Y\mathbf{A}\mathbf{x}
       - 2\mathbf{x}^T\mathbf{\Sigma}^{-1}_X\bm{\mu}_X \nonumber \\
      &- 2\mathbf{x}^T\mathbf{A}^T\mathbf{\Sigma}^{-1}_Y\mathbf{y} + c_1 \nonumber \\
    = & (\mathbf{x}-\bm{\mu})^T \mathbf{\Sigma}^{-1} (\mathbf{x}-\bm{\mu}) + c_2 \nonumber
\end{align}
where
\begin{align}
    \mathbf{\Sigma}^{-1} & = \mathbf{\Sigma}^{-1}_X + \mathbf{A}^T\mathbf{\Sigma}^{-1}_Y\mathbf{A} \nonumber \\
    \bm{\mu} & = \mathbf{\Sigma}(\mathbf{\Sigma}^{-1}_X\bm{\mu}_X + \mathbf{A}^T\mathbf{\Sigma}^{-1}_Y\mathbf{y})  \nonumber \\
    c_1 & = \bm{\mu}^T_X\mathbf{\Sigma}^{-1}_X\bm{\mu}_X + \mathbf{y}^T\mathbf{\Sigma}^{-1}_Y\mathbf{y} \nonumber \\
    c_2 & = -\bm{\mu}^T\mathbf{\Sigma}^{-1}\bm{\mu} + c_1 \nonumber
\end{align}
Therefore, the product of two Gaussian is:
\begin{align}
      & \mathcal{N}(\mathbf{x}; \bm{\mu}_X, \mathbf{\Sigma}_X)
        \times \mathcal{N}(\mathbf{y};\bm{\mu}_Y, \mathbf{\Sigma}_Y)  \nonumber \\
    = & \frac{1}{\sqrt{(2\pi)^{d_Y} \det(\mathbf{\Sigma}_Y) }}  \sqrt{\frac{\det(\mathbf{\Sigma})}{\det(\mathbf{\Sigma}_X)}}
        \exp\left\{-\frac{1}{2}c_2\right\} \nonumber \\
      & \times \frac{1}{\sqrt{(2\pi)^{d_X} \det(\mathbf{\Sigma}) }}
      \exp\left\{-\frac{1}{2} (\mathbf{x}-\bm{\mu})^T \mathbf{\Sigma}^{-1} (\mathbf{x}-\bm{\mu}) \right\} \nonumber \\
   = & c_3 \cdot \mathcal{N}(\mathbf{x}; \bm{\mu}, \mathbf{\Sigma}) \nonumber
\end{align}
where
\begin{equation}
    c_3 = \frac{e^{-c_2/2}}{\sqrt{(2\pi)^{d_Y}}} \sqrt{\frac{\det(\mathbf{\Sigma})}{\det(\mathbf{\Sigma}_X)\det(\mathbf{\Sigma}_Y)}}
            \nonumber
\end{equation}

The third term in the integral of Eq.\eqref{eq:int_norm_logit}, defined in Eq.\eqref{eq:pD_trans_disc},
can be re-written as:
\begin{align}
    & p(D_t^{d+1}|D_{t-1}^d, S_{t-1}^k, Z^l_{t-1}, X^{\mathbf{x}}_{t-1}) \nonumber \\
  = & \frac{1}{1+e^{\nu_k (d-\beta_k) + \bm{\omega}_{kl}^T \mathbf{x}}} \nonumber \\
  = & \frac{1}{1+e^{-\left[\beta + \bm{\omega}^T \mathbf{x} \right]}} \nonumber \\
  = & \mathcal{F}\left(\vec{\bm{\omega}}^T \mathbf{x} + \frac{\beta}{||\bm{\omega}||}; \frac{1}{||\bm{\omega}||} \right) \nonumber
\end{align}
where $\beta=-\nu_k (d-\beta_k)$, $\bm{\omega}=-\bm{\omega}_{kl} = ||\bm{\omega}||\cdot\vec{\bm{\omega}}$,
and $\mathcal{F}(x;\alpha)=\frac{1}{1+e^{-x/\alpha}}$ is logistic (or Fermi) function.
The probability for $p(D_t^{1}|\cdot)$ can be obtained accordingly.

To convert the integral in Eq.\eqref{eq:int_norm_logit} into a single variable integral,
we further introduce a linear transformation:
\begin{equation}
    \mathbf{v} = \mathbf{W}^T \mathbf{x} \nonumber
\end{equation}
where $\mathbf{W}^T\mathbf{W}=\mathbf{I}$ is orthonormal, and $\mathbf{W}(:, 1)=\bm{\vec{\omega}}$.
For Gaussian variable, we have:
\begin{align}
            & (\mathbf{x}-\bm{\mu})^T \mathbf{\Sigma}^{-1} (\mathbf{x}-\bm{\mu}) \nonumber \\
    = & (\mathbf{W}\mathbf{v}-\bm{\mu})^T \mathbf{\Sigma}^{-1} (\mathbf{W}\mathbf{v}-\bm{\mu}) \nonumber \\
    = & (\mathbf{v}-\mathbf{W}^T\bm{\mu})^T \mathbf{W}^T\mathbf{\Sigma}^{-1}\mathbf{W} (\mathbf{v}-\mathbf{W}^T\bm{\mu}) \nonumber
\end{align}
Therefore,
\begin{equation}
    \mathcal{N}(\mathbf{x}; \bm{\mu}, \mathbf{\Sigma}) = 
    \mathcal{N}(\mathbf{v}; \mathbf{W}^T\bm{\mu}, \mathbf{W}^T\mathbf{\Sigma}\mathbf{W}) \nonumber
\end{equation}

Now we are ready to evaluate Eq.\eqref{eq:int_norm_logit} as:
\begin{align}
    & c_3 \int \mathcal{N}(\mathbf{x}; \bm{\mu}, \mathbf{\Sigma}) 
            \mathcal{F}\left( \bm{\vec{\omega}}^T \mathbf{x} + \beta/||\bm{\omega}||; 1/||\bm{\omega}|| \right) 
            d\mathbf{x} \nonumber \\
  = & \frac{c_3}{|\det(\mathbf{W}^T)|} \int \mathcal{N}(\mathbf{v}; \mathbf{W}^T\bm{\mu}, \mathbf{W}^T\mathbf{\Sigma}\mathbf{W})
            \nonumber \\
    &       \;\;\;\;\;\;\;\;\;\;\;\;\;\;\;\;\;\;\; \times \mathcal{F}\left( v_1 + \beta/||\bm{\omega}||; 1/||\bm{\omega}|| \right) 
            d\mathbf{v} \nonumber \\
  = & c_3 \int \mathcal{N}(v_1; \bm{\vec{\omega}}^T\bm{\mu}, \bm{\vec{\omega}}^T\mathbf{\Sigma}\bm{\vec{\omega}})
            \mathcal{F}\left( v_1 + \beta/||\bm{\omega}||; 1/||\bm{\omega}|| \right) d{v_1} \nonumber \\
  = & c_3 \int \mathcal{N}(v; 0, \bm{\vec{\omega}}^T\mathbf{\Sigma}\bm{\vec{\omega}})
            \mathcal{F}\left( v + \beta/||\bm{\omega}|| + \bm{\vec{\omega}}^T\bm{\mu}; 1/||\bm{\omega}|| \right) d{v} \nonumber \\
  \approx & c_3 \cdot \mathcal{F}\left( \frac{1}{||\bm{\omega}||}(\beta + \bm{\omega}^T\bm{\mu}); 
                                \sqrt{1+\frac{\pi}{8} \bm{\omega}^T\mathbf{\Sigma}\bm{\omega}} \right) \nonumber
\end{align}
where the approximation follows from \cite{maragakis08}.

{\small
\bibliographystyle{ieee}
\bibliography{./reference}

\begin{thebibliography}{10}\itemsep=-1pt

\bibitem{Arulampalam02}
M.~Arulampalam, S.~Maskell, N.~Gordon, and T.~Clapp.
\newblock A tutorial on particle filters for on-line nonlinear/non-{G}aussian
  {B}ayesian tracking.
\newblock {\em IEEE Trans. on Signal Processing}, 50(2):174--188, 2002.

\bibitem{Bakis:Continuous}
R.~Bakis.
\newblock Continuous speech word recognition via centisecond acoustic states.
\newblock {\em unpublished paper presented at the meeting of the Acoustics
  Society of America}, 1976.

\bibitem{Barber:Expectation}
D.~Barber.
\newblock Expectation correction for smoothed inference in switching linear
  dynamical systems.
\newblock {\em Journal of Machine Learning Research}, pages 2515--2540, 2006.

\bibitem{Bicego:Sparseness}
M.~Bicego, M.~Cristani, and V.~Murino.
\newblock Sparseness achievement in hidden {M}arkov models.
\newblock {\em Proc.of ICIAP'07}, pages 67--72, 2007.

\bibitem{Cemgil:generative}
A.~Cemgil, H.~Kappen, and D.~Barber.
\newblock A generative model for music transcription.
\newblock In {\em IEEE Transactions on Audio, Speech, and Language Processing},
  pages 679--694, 2006.

\bibitem{Chib:Non}
S.~Chib and M.~J. Dueker.
\newblock Non-{M}arkovian regime switching with endogenous states and
  time-varying state strengths.
\newblock Econometric Society 2004 North American Summer Meetings 600,
  Econometric Society, 2004.

\bibitem{Cutler:Robust}
R.~Cutler and L.~Davis.
\newblock Robust real-time periodic motion detection, analysis, and
  applications.
\newblock {\em IEEE Trans. on Pattern Analysis and Machine Intelligence},
  22(8):781--796, 2000.

\bibitem{Doucet00}
A.~Doucet, N.~d. Freitas, K.~P. Murphy, and S.~J. Russell.
\newblock Rao-{B}lackwellised particle filtering for dynamic {B}ayesian
  networks.
\newblock In {\em Proceedings of the 16th Conference on Uncertainty in
  Artificial Intelligence}, pages 176--183, 2000.

\bibitem{Ferguson:Variable}
J.~Ferguson.
\newblock Variable duration models for speech.
\newblock {\em Symp. Application of Hidden Markov Models to Text and Speech,
  Institute for Defense Analyses, Princeton, NJ}, pages 143--179, 1980.

\bibitem{Fox:Nonparametric}
E.~Fox, E.~Sudderth, M.~Jordan, and A.~Willsky.
\newblock Nonparametric {B}ayesian learning of switching linear dynamical
  systems.
\newblock {\em Proc. of NIPS'09}, 2009.

\bibitem{Ghahramani:Factorial}
Z.~Ghahramani and M.~Jordan.
\newblock Factorial hidden {M}arkov models.
\newblock {\em Proc.of NIPS}, 1996.

\bibitem{Harchaoui:Kernel}
Z.~Harchaoui, F.~Bach, and E.~Moulines.
\newblock Kernel changepoint analysis.
\newblock {\em Proc. of NIPS'09}, 2009.

\bibitem{Hoai:Joint}
M.~Hoai, Z.~Lan, and F.~Torre.
\newblock Joint segmentation and classification of human action in video.
\newblock {\em Proc. of CVPR'11}, 2011.

\bibitem{Hoffken:Switching}
M.~Hoffken, D.~Oberhoff, and M.~Kolesnik.
\newblock Switching hidden {M}arkov models for learning of motion patterns in
  videos.
\newblock {\em Proc. of ICANN'09}, pages 757--766, 2009.

\bibitem{Julier:new}
S.~Julier and J.~Uhlmann.
\newblock A new extension of the {K}alman filter to nonlinear systems.
\newblock {\em Proc of AeroSense: The 11th International Symposium on
  Aerospace/Defence Sensing, Simulation and Control}, pages 182--193, 1997.

\bibitem{Khan04}
Z.~Khan, T.~Balch, and F.~Dellaert.
\newblock A {R}ao-{B}lackwellized particle filter for {E}igen{T}racking.
\newblock In {\em Proceedings of the IEEE computer society conference on
  Computer vision and pattern recognition}, pages 980--987, 2004.

\bibitem{Kjellstrom:Simultaneous}
H.~Kjellstrom, J.~Romero, D.~Martinez, and D.~Kragic.
\newblock Simultaneous visual recognition of manipulation actions and
  manipulated objects.
\newblock {\em Proc. of ECCV'08}, 2008.

\bibitem{Lafferty:Conditional}
J.~Lafferty, A.~McCallum, and F.~Pereira.
\newblock Conditional random fields: probabilistic models for segmenting and
  labeling sequence data.
\newblock {\em Proc. of ICML'01}, 2001.

\bibitem{Lazebnik:Beyond}
S.~Lazebnik, C.~Schmid, and J.~Ponce.
\newblock Beyond bags of features: spatial pyramid matching for recognizing
  natural scene categories.
\newblock {\em Proc. of CVPR'06}, 2006.

\bibitem{Lerner:inferencein}
U.~Lerner and R.~Parr.
\newblock Inference in hybrid networks: theoretical limits and practical
  algorithms.
\newblock In {\em In UAI}, pages 310--318, 2001.

\bibitem{Levinson:Continuously}
S.~Levinson.
\newblock Continuously variable duration hidden {M}arkov models for automatic
  speech recognition.
\newblock {\em Comput. Speech Lang.}, pages 29--45, 1986.

\bibitem{Lipovetsky:Double}
S.~Lipovetsky.
\newblock Double logistic curve in regression modeling.
\newblock {\em Journal of Applied Statistics}, pages 1785--793, 2010.

\bibitem{Liu:Recognizing}
J.~Liu, B.~Kuipers, and S.~Savarese.
\newblock Recognizing human actions by attributes.
\newblock {\em Proc. of CVPR'11}, 2011.

\bibitem{maragakis08}
P.~Maragakis, F.~Ritort, C.~Bustamante, M.~Karplus, and G.~E. Crooks.
\newblock Bayesian estimates of free energies from nonequilibrium work data in
  the presence of instrument noise.
\newblock {\em Journal of Chemical Physics}, 129, 2008.

\bibitem{McCallum:Maximum}
A.~McCallum, D.~Freitag, and F.~Pereira.
\newblock Maximum entropy {M}arkov models for information extraction and
  segmentation.
\newblock {\em Proc. of ICML'00}, pages 591--598, 2000.

\bibitem{Morency:Latent}
L.~Morency, A.~Quattoni, and T.~Darrell.
\newblock Latent-dynamic discriminative models for continuous gesture
  recognition.
\newblock {\em Proc. of CVPR'07}, 2007.

\bibitem{Ning:Conditional}
H.~Ning, W.~Xu, Y.~Gong, and T.~Huan.
\newblock Latent pose estimator for continuous action recognition.
\newblock {\em Proc. of ECCV'08}, 2008.

\bibitem{Oh:Learning}
S.~Oh, J.~Rehg, T.~Balch, and F.~Dellaert.
\newblock Learning and inference in parametric switching linear dynamic
  systems.
\newblock {\em Proc. of ICCV'05}, 2005.

\bibitem{Sang:Parameterized}
S.~M. Oh, J.~M. Rehg, and F.~Dellaert.
\newblock Parameterized duration modeling for switching linear dynamic systems.
\newblock {\em Proc. of CVPR'06}, pages 1--8, 2006.

\bibitem{Ozay:Sequential}
N.~Ozay, M.~Sznaier, and C.~O.
\newblock Sequential sparsificarion for change detection.
\newblock {\em Proc. of CVPR'08}, 2008.

\bibitem{Raptis:Spike}
M.~Raptis, K.~Wnuk, and S.~Soatto.
\newblock Spike train driven dynamical models for human actions.
\newblock {\em Proc. of CVPR'10}, pages 2077--2084, 2010.

\bibitem{Satkin:Modeling}
S.~Satkin and M.~Hebert.
\newblock Modeling the temporal extent of actions.
\newblock {\em Proc. of ECCV'10}, 2010.

\bibitem{Sminchisescu:Conditional}
C.~Sminchisescu, A.~Kanaujia, Z.~Li, and D.~Metaxas.
\newblock Conditional models for contextual human motion recognition.
\newblock {\em Proc. of ICCV'05}, 2005.

\bibitem{Sung:Human}
J.~Sung, C.~Ponce, B.~Selman, and A.~Saxena.
\newblock Human activity detection from {R}{G}{B}{D} images.
\newblock {\em Proc. of AAAI'11}, 2011.

\bibitem{Jinjun:LLC}
J.~Wang, J.~Yang, F.~Lv, and K.~Yu.
\newblock Locality-constrained linear coding for image classification.
\newblock {\em Proc. of CVPR'10}, 2010.

\bibitem{Wang:Event}
Z.~Wang, E.~Kuruoglu, X.~Yang, Y.~Xu, and S.~Yu.
\newblock Event recognition with time varying hidden {M}arkov model.
\newblock {\em Proc.of ICASSP'09}, pages 1761--1764, 2009.

\bibitem{Weinland:Free}
D.~Weinland, R.~Ronfard, and E.~Boyer.
\newblock Free viewpoint action recognition using motion history volumes.
\newblock {\em Computer Vision and Image Understanding}, 2006.

\bibitem{Yoshimura:Duration}
T.~Yoshimura, K.~Tokuda, T.~Masuko, T.~Kobayashi, and T.~Kitamura.
\newblock Duration modeling for {H}{M}{M}-based speech synthesis.
\newblock {\em Proc. of ICSLP'98}, 1998.

\bibitem{Yu:Hidden}
S.-Z. Yu.
\newblock Hidden semi-{M}arkov models.
\newblock {\em Artificial Intelligence}, pages 215--243, 2010.

\bibitem{Zhou:Aligned}
F.~Zhou, F.~Torre, and J.~K. Hodgins.
\newblock Aligned cluster analysis for temporal segmentation of human motion.
\newblock {\em Proc. of IEEE Conference on Automatic Face and Gestures
  Recognition}, 2008.

\end{thebibliography}
}

\end{document}